\documentclass[10pt,twocolumn,letterpaper]{article}

\usepackage{cvpr}              

\usepackage[dvipsnames]{xcolor}

\usepackage{multirow}

\newcommand{\Lagr}{\mathcal{L}}

\usepackage{algorithm,algcompatible,amsmath,}
\usepackage{makecell}
\algdef{SE}[SUBALG]{Indent}{EndIndent}{}{\algorithmicend\ }%
\algnewcommand\INPUT{\item[\textbf{Input:}]}%
\algnewcommand\OUTPUT{\item[\textbf{Output:}]}%
\algnewcommand\PARAM{\item[\textbf{Optimizable Parameter:}]}%

\algdef{SE}[SUBALG]{Indent}{EndIndent}{}{\algorithmicend\ }%
\algtext*{Indent}
\algtext*{EndIndent}

\definecolor{cvprblue}{rgb}{0.21,0.49,0.74}
\usepackage[pagebackref,breaklinks,colorlinks,citecolor=cvprblue]{hyperref}

\def\paperID{8689} 
\def\confName{CVPR}
\def\confYear{2024}

\def\titlePrefix{NARUTO}
\title{
\titlePrefix: Neural Active Reconstruction from Uncertain Target Observations
}

\author{
Ziyue Feng 
    \thanks{Equal contribution} 
    \thanks{Work done as an intern at OPPO US Research Center} 
    \textsuperscript{,1,2}
    \hspace{15pt}
Huangying Zhan 
    \footnotemark[1]
    \thanks{Corresponding author (zhanhuangying.work@gmail.com)}
    \textsuperscript{,1} 
    \hspace{15pt}
Zheng Chen
    \footnotemark[2]
    \textsuperscript{,1,3} 
    \hspace{15pt}
Qingan Yan
    \textsuperscript{1} 
    \\
Xiangyu Xu
    \textsuperscript{1} 
    \hspace{10pt}
Changjiang Cai
    \textsuperscript{1} 
    \hspace{10pt}
Bing Li
    \textsuperscript{2} 
    \hspace{10pt}
Qilun Zhu
    \textsuperscript{2} 
    \hspace{10pt} 
Yi Xu
    \textsuperscript{1} 
    \and 
\textsuperscript{1}
    OPPO US Research Center \and
\textsuperscript{2}
    Clemson University \and
\textsuperscript{3}
    Indiana University
}

\begin{document}
\maketitle
\begin{abstract}
We present \titlePrefix{}, a neural active reconstruction system that combines a hybrid neural representation with uncertainty learning, enabling high-fidelity surface reconstruction. 
Our approach leverages a multi-resolution hash-grid as the mapping backbone, chosen for its exceptional convergence speed and capacity to capture high-frequency local features.
The centerpiece of our work is the incorporation of an uncertainty learning module that dynamically quantifies reconstruction uncertainty while actively reconstructing the environment. 
By harnessing learned uncertainty, we propose a novel uncertainty aggregation strategy for goal searching and efficient path planning. 
Our system autonomously explores by targeting uncertain observations and reconstructs environments with remarkable completeness and fidelity. 
We also demonstrate the utility of this uncertainty-aware approach by enhancing SOTA neural SLAM systems through an active ray sampling strategy.
Extensive evaluations of \titlePrefix{} in various environments, using an indoor scene simulator, confirm its superior performance and state-of-the-art status in active reconstruction, as evidenced by its impressive results on benchmark datasets like Replica and MP3D. 
Project page: 
\href{https://oppo-us-research.github.io/NARUTO-website/}{oppo-us-research.github.io/NARUTO-website/}
\end{abstract}    
\vspace{-15pt}
\section{Introduction}
\label{sec:intro}

\begin{figure}[t!]
        \centering
		\includegraphics[width=1.0\columnwidth]    {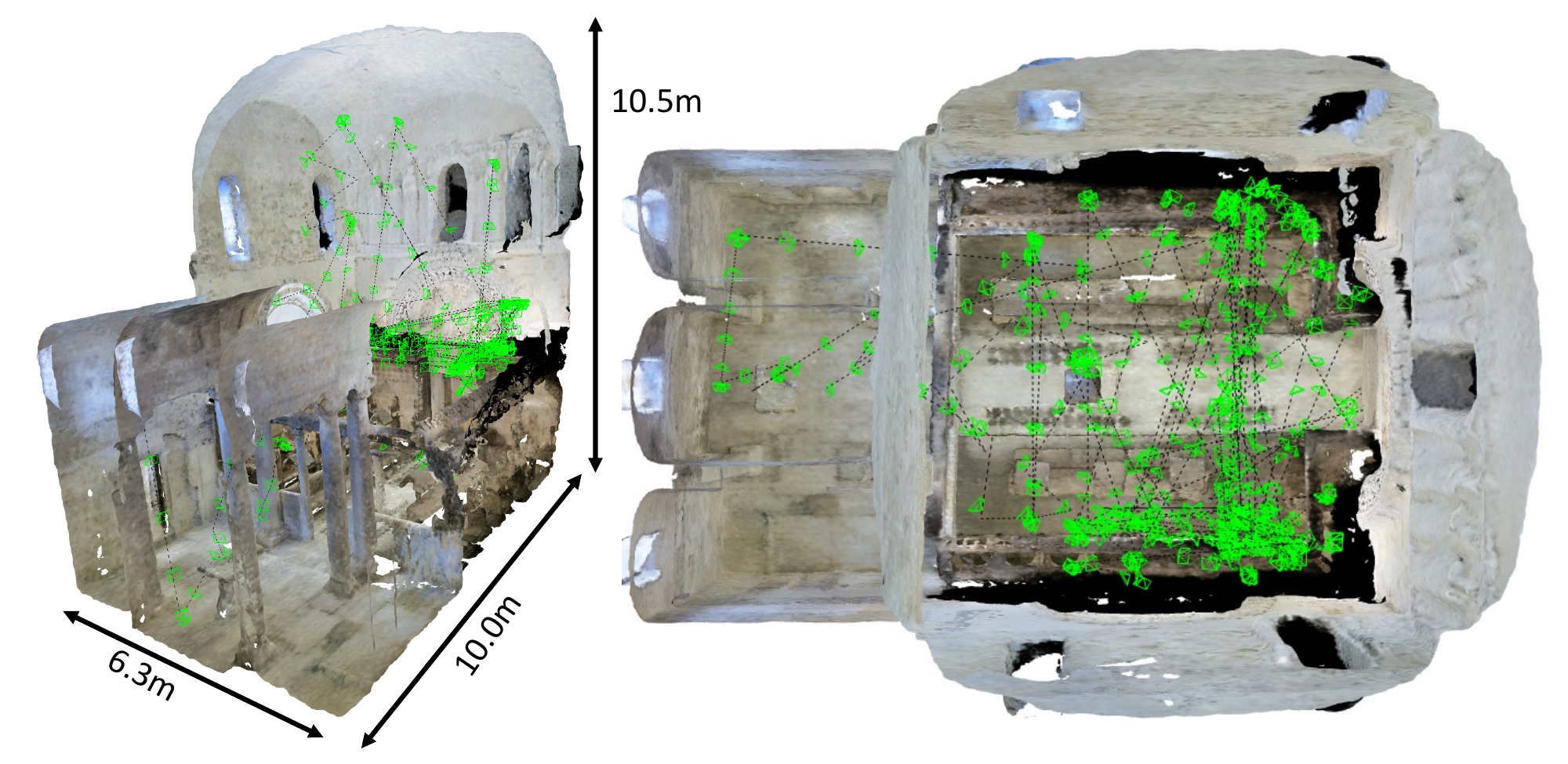}
		\caption{
            We introduce a neural active reconstruction system, named \textit{NARUTO}, which is guided by learned uncertainty. 
            NARUTO enables an agent to identify areas of uncertainty and proactively investigate these regions to minimize reconstruction ambiguity. Consequently, this approach facilitates the incremental completion of the entire scene's reconstruction. 
            \textit{NARUTO} represents the first neural active Reconstruction system capable of functioning in large-scale environments with unrestricted movement.
            }
		\label{fig:teaser}
		\vspace{-15pt}
\end{figure}
In the realm of computer vision research, one of the most notable advancements is the ability to generate detailed 3D reconstructions from an array of 2D images or scene videos.
This intricate process, executed in real-time, involves progressive 3D modeling as additional visual data is assimilated, predominantly through the use of Simultaneous Localization and Mapping (SLAM). 
In many robotic applications, SLAM systems are instrumental for tasks such as planning and navigation. 
This integration of localization, mapping, planning, and navigation tasks forms the essence of what is known as Active SLAM.
Our paper specifically addresses a subset of Active SLAM, termed Active Reconstruction, under the assumption that localization is already established. 
We venture into an innovative exploration of Active Reconstruction by adopting a sophisticated, learned hybrid neural representation 
. 
In this work, we devise methodologies capable of meticulously planning and maneuvering camera trajectories to enhance the completeness and quality of the scene's reconstruction.

Neural representations, particularly implicit Neural Radiance Fields (NeRFs), have recently been applied in diverse geometric applications, such as 
3D object reconstruction \cite{park2019deepsdf}, 
novel view rendering \cite{mildenhall2021nerf, zhang2020nerf++, yu2021pixelnerf, pumarola2021d}, 
surface reconstruction \cite{li2022bnv, azinovic2022neural}, 
and generative models \cite{schwarz2020graf, niemeyer2021giraffe}.
While many of these methods focus on posed cameras, recent efforts have expanded to broader tasks like structure from motion \cite{wang2021nerf--, lin2021barf, chng2022gaussian} and SLAM \cite{sucar2021imap, zhu2022niceslam, zhu2023nicer, wang2023coslam}.
Despite the impressive capabilities of NeRFs, their processing speed remains a challenge.
To address this, more efficient hybrid neural representations have been developed \cite{SunSC22dvgo, mueller2022instant}.

Integrating these representations into active vision applications continues to pose significant challenges.
Existing research utilizing neural representations for path planning is limited \cite{adamkiewicz2022vision}, and only a handful of recent studies have explored active reconstruction with neural representations \cite{ran2022neurar, lee2022uncertainty, pan2022activenerf, zhan2022activermap, yan2023active}.
These approaches, while innovative, often suffer from the inherent slow speeds of NeRFs \cite{ran2022neurar, lee2022uncertainty, pan2022activenerf}.
Moreover, they typically constrain the movement of agents to a lower degree-of-freedom (DoF) within restricted areas, such as specific locations \cite{pan2022activenerf, lee2022uncertainty}, within a hemisphere \cite{ran2022neurar, zhan2022activermap}, or on a 2D plane \cite{yan2023active}.

To overcome the aforementioned limitations, we introduce \textit{\titlePrefix{}}, a groundbreaking neural active reconstruction system. 
\textit{\titlePrefix{}} unites a hybrid neural representation with a novel uncertainty-aware planning module, excelling in high-fidelity surface reconstruction and proactive planning, shown in \cref{fig:teaser}. Our key contributions are as follows:
\begin{itemize}
    \setlength{\itemsep}{0pt}
    \setlength{\parskip}{0pt}
    \setlength{\parsep}{0pt}
    \item The \textit{first} neural active reconstruction system operating with 6DoF movement in unrestricted spaces.
    \item An uncertainty learning module quantifies reconstruction uncertainty in real-time.
    \item A novel uncertainty-aware planning features a meticulously designed uncertainty aggregation for goal searching, and efficient path planning.
    \item Active ray sampling strategy enhances the performance and stability of mapping modules across various tasks.
    \item Achieving exceptional active reconstruction performance, advancing state-of-the-art in reconstruction completeness from 73\% to 90\%.
\end{itemize}

\section{Related Work} 
\label{sec:rel_work}

\paragraph{Active Reconstruction}
In autonomous robotics, essential capabilities include localization, mapping, planning, and motion control \cite{siegwart2011introduction}. 
These elements have led to research areas like 
visual odometry \cite{scaramuzza2011visual, zhan2020visual}, 
monocular depth estimation \cite{eigen2014depth, zhan2018unsupervised, bian2019unsupervised, feng2022disentangling, feng2022advancing},
multi-view stereo \cite{sun2003stereo, hirschmuller2005accurate, seitz2006comparison, yao2018mvsnet, liu2022planemvs, cai2023riav, chen2023learning}, 
structure-from-motion (SfM) \cite{schonberger2016structure}, 
path planning \cite{Hart1968, lavalle2001rrt, kuffner2000rrtconnect, feng2018model}, 
and SLAM \cite{thrun2002probabilistic, durrant2006simultaneous, davison2007monoslam, cadena2016past, wang2023edi}. 
Active SLAM, which combines these approaches for autonomous localization, mapping, and planning, minimizes uncertainties in environmental modeling \cite{davison2002simultaneous}. 
We refer readers to the survey papers \cite{cadena2016past, lluvia2021active, placed2022survey} for a comprehensive discussion regarding the development of active SLAM.
Our focus is on active reconstruction, often investigated as exploration problems \cite{thrun1991active, feder1999adaptive, bourgault2002information, makarenko2002experiment, stachniss2004exploration, newman2003autonomous, stachniss2009robotic}.
a problem that seeks optimal movements for accurate environmental representations \cite{connolly1985determination}, primarily for scene and object reconstruction from multiple viewpoints \cite{maver1993occlusions, pito1999solution, kriegel2015efficient, isler2016information, delmerico2018comparison, peralta2020next}.

\vspace{-11pt}

\paragraph{Neural Representaitons}
NeRFs \cite{mildenhall2021nerf} use multi-layer perceptrons (MLPs) to represent scenes as continuous neural radiance fields. 
NeRF's potential has been demonstrated in a range of applications, from novel view rendering \cite{mildenhall2021nerf, zhang2020nerf++, yu2021pixelnerf, pumarola2021d} to object \cite{park2019deepsdf, mildenhall2021nerf} and surface reconstruction \cite{li2022bnv, azinovic2022neural}, as well as in generative models \cite{schwarz2020graf, niemeyer2021giraffe}, Structure-from-Motion \cite{wang2021nerf--, lin2021barf, chng2022gaussian}.
NeRFs are trained by comparing rendered images with accurately posed ones. 
However, the volume rendering process \cite{kajiya1984ray}, which involves querying numerous sample points for image rendering, makes training NeRFs time-intensive, often requiring about a day for simple scenes. 
While efforts have been made to accelerate NeRFs \cite{Reiser2021ICCV, deng2022depth, liu2021neulf, chen2023neurbf}, these methods still fall short of real-time application speeds. 
Recent work \cite{SunSC22dvgo, yu_and_fridovichkeil2021plenoxels, mueller2022instant, Chen2022tensorf} have achieved fast speed through hybrid representations, combining implicit and explicit elements for light and density fields, respectively.
The advancement in hybrid representations has been instrumental in meeting the real-time requirements of SLAM challenges \cite{zhu2022niceslam, zhu2023nicer, wang2023coslam}.
Despite these advancements, applying neural representations in active vision problems is still an underexplored area.

\vspace{-12pt}
\paragraph{Neural Active Vision}
Our research builds upon prior works that have explored the use of NeRFs for path planning \cite{adamkiewicz2022vision} and active reconstruction \cite{ran2022neurar, lee2022uncertainty, pan2022activenerf}. 
\cite{adamkiewicz2022vision} derives optimal paths for navigation from the NeRF-based scene representation. 
Recent studies \cite{lee2022uncertainty, pan2022activenerf, ran2022neurar} have focused on active mapping, optimizing NeRFs with next-best-view selection strategies. 
However, these approaches are constrained by the inherent slow speed of NeRFs, limiting their real-time application in robotics. 
\cite{zhan2022activermap} proposes an efficient framework using hybrid representations to address these speed concerns. Meanwhile, works like \cite{chaplot2020activeneuralslam, georgakis2022uncertainty, yan2023active} have expanded the scope from object-centric reconstruction \cite{lee2022uncertainty, pan2022activenerf, ran2022neurar, zhan2022activermap} to larger indoor environments.
However, these methods still restrict camera motion to a hemisphere or a 2D plane. 
In contrast, \textit{\titlePrefix{}} enables 6DoF exploration in unrestricted spaces.
\section{\titlePrefix{}: Neural Active Reconstruction}
\label{sec:method}

\begin{figure*}[th!]
        \centering
		\includegraphics[width=1.0\textwidth]{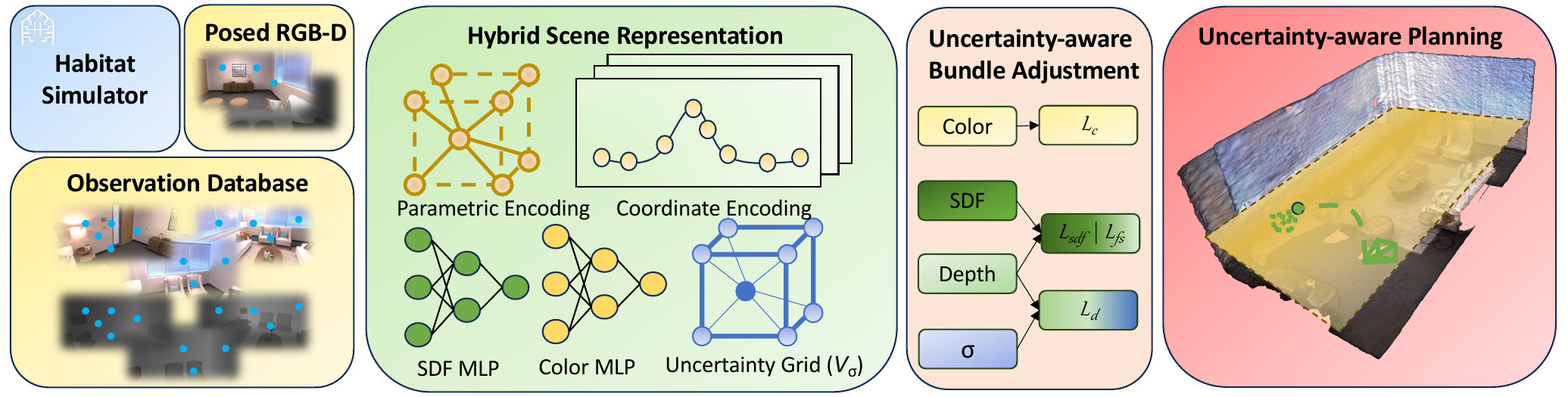}
		\caption{
                \textbf{NARUTO framework}
                Upon reaching a keyframe step, HabitatSim \cite{savva2019habitat} generates posed RGB-D images.
                A select number of pixels from these images are sampled and stored in the observation database.
                Utilizing a mixed ray sampling strategy (combining Random and Active methods), a subset of rays is selected from the current keyframe and the database.
                These rays are then processed through the Hybrid Scene Representation (Map) to deduce the corresponding color, Signed Distance Function (SDF), depth, and uncertainty values.
                The predictions derived from this process facilitate uncertainty-aware bundle adjustment, updating both the scene's geometry and reconstruction uncertainty.
                Subsequently, the Map is refreshed, and our novel uncertainty-aware planning algorithm is employed to determine a goal and trajectory based on the SDFs and uncertainties.
                The agent then executes the planned action.
                }
		\label{fig:framework}
		\vspace{-15pt}
\end{figure*}

In this section, we introduce \textit{\titlePrefix{}} (\cref{fig:framework}), a pioneering neural framework in active reconstruction with uncertainty-aware planning.
Our approach begins with the neural 3D mapping module, utilizing a hybrid representation for real-time, high-fidelity surface reconstruction.
We incorporate Co-SLAM \cite{wang2023coslam} as the mapping backbone, as discussed in \cref{sec:method:neural_3d_mapping}, laying the groundwork for 3D reconstruction using hybrid neural representation.
Building upon this, \cref{sec:method:uncertainty} delves into the framework's core, illustrating the joint optimization method that fuses bundle adjustment with uncertainty learning.
In \cref{sec:method:active_plan}, we present the uncertainty-aware planning module for goal searching and path planning.
\cref{sec:method:active_ray} introduces a versatile active ray sampling module.
This module, leveraging the learned uncertainty, is designed for seamless integration into existing neural mapping methodologies.
Concluding this section, we summarize the procedure of active reconstruction in \cref{sec:method:act_recon}. 

\subsection{Neural 3D Mapping}
\label{sec:method:neural_3d_mapping}
\paragraph{\textbf{Implicit Neural Mapping}}
Recent advancements have established neural implicit representations as notably expressive and compact, effectively encoding scenes' appearance and 3D geometry.
A series of prior works, including \cite{sucar2021imap, zhu2022niceslam, li2022bnv, zhu2023nicer, wang2023coslam}, have demonstrated the applicability of neural representation in 3D reconstruction.
Given a stream of RGB-D images, dense mapping with representations, such as radiance fields and truncated signed distance fields (TSDF), can be achieved by optimizing a neural representation via rendering supervision.
TSDF, in particular, is widely used for neural surface reconstruction.
Coordinate-based neural representations are often employed to map world coordinates $\textbf{x}$ to color $\textbf{c}$ and TSDF value $s$.
%
%

\vspace{-10pt}

\paragraph{\textbf{Hybrid Representation}}
MLPs are widely utilized as coordinate-based implicit representations for high-fidelity scene reconstruction, owing to their coherence and smoothness.
However, they are not without drawbacks, such as slow convergence and catastrophic forgetting in continual learning scenarios, as identified in \cite{yan2021continual, cai2023clnerf}.
To address these challenges, we apply several innovative solutions introduced by Co-SLAM \cite{wang2023coslam}.
Among these is a joint coordinate and parametric encoding, designed to enhance fidelity while expediting training processes.
The incorporation of one-blob coordinate encoding $\gamma(\textbf{x})$ \cite{muller2019neural} with a multi-resolution hash-based feature grid achieves rapid querying speeds, efficient memory usage, and a notable hole-filling capability.
In this setup, the feature vector $V_{\alpha}(\textbf{x})$ at each sampled point $\textbf{x}$ is obtained through trilinear interpolation on the feature grid.
The geometry decoder $f_{\tau}$ predicts an SDF value $s$ and a feature vector $\textbf{h}$.
Additionally, the color MLP, denoted as $f_{\phi}$, calculates the color value.
\begin{align}
f_{\tau}(\gamma(\textbf{x}), V_{\alpha}(\textbf{x})) \mapsto (\textbf{h}, s) 
\text{ ; }
f_{\phi}(\gamma(\textbf{x}), \textbf{h}) \mapsto \textbf{c},
\end{align}
where $\{\alpha, \phi, \tau\}$ represents the learnable parameters that can be optimized in the bundle adjustment.

\paragraph{\textbf{Bundle Adjustment}}
Bundle Adjustment (BA) in neural SLAM typically employs volumetric rendering optimization \cite{sucar2021imap, zhu2022niceslam, wang2023coslam}.
Instead of storing full images, we execute BA on sparse samples from the keyframes, enabling more frequent keyframe insertions and a larger keyframe collection.
For this process, given a camera origin $\textbf{o}$ and a ray direction $\textbf{r}$, 3D points are sampled along the ray, based on predefined depths $d_i$: $\textbf{x}_i = \textbf{o} + d_i\textbf{r}$.
The color $\hat{\textbf{c}}$ and depth $\hat{\textbf{d}}$ can be rendered:
\begin{align}
    \hat{\textbf{c}} = \frac{1}{\sum_{i=1}^{M} w_i} \sum_{i=1}^{M} w_i \textbf{c}_i 
    \text{ , }
    \hat{d} = \frac{1}{\sum_{i=1}^{M} w_i} \sum_{i=1}^{M} w_i d_i,
\end{align}
where $w_i = \varphi(\frac{s_i}{tr})\varphi(-\frac{s_i}{tr})$ represents the weights computed along the ray, obtained by applying Sigmoid functions $\varphi(.)$ to the predicted SDF $\textit{s}_i$ within a truncated range $tr = 10\text{cm}$.

Post rendering, a multi-objective function is minimized to execute bundle adjustment, incorporating color and depth rendering losses.
These losses are calculated between the rendered results ($\hat{\textbf{c}}$, $\hat{d}$) and the observed values ($\textbf{c}^o$,$D$):
\begin{align}
    \Lagr_{c} = \frac{1}{N} \sum_{i=1}^{N}(\hat{\textbf{c}}_i - \textbf{c}_{i}^o)^2
    \text{ , }
    \Lagr_{d} = \frac{1}{|R_d|} \sum_{r \in R_d}(\hat{d}_r - D_r)^2
    \label{eqn:depth_loss}
\end{align}
where $N=2148$, $R_d$ denotes the set of rays with valid depths, and $D_r$ corresponds to the pixel on the image plane.

Following \cite{wang2023coslam}, we apply additional regularizations to enhance reconstruction quality.
For samples within the truncation region $S_r^{tr}$, SDF loss is approximated by the distance between the sampled point and its observed depth value.
Conversely, 
for points outside the truncation region $S_r^{fs}$, 
a free-space loss ensures SDF predictions equal to $tr$:
\begin{align}
    \Lagr_{sdf} &= \frac{1}{|R_d|} \sum_{r \in R_d} \frac{1}{|S_r^{tr}|} 
    \sum_{p \in S_r^{tr}}(s_p - (D_p-d))^2 \\
    \Lagr_{fs} &= \frac{1}{|R_d|} \sum_{r \in R_d} \frac{1}{|S_r^{fs}|} 
    \sum_{p \in S_r^{fs}}(s_p - tr)^2
    .
\end{align}
To ensure smooth reconstructions in unobserved free-space regions, we apply a feature smoothness regularization on the interpolated features $V_\alpha(\textbf{x})$:
\begin{align}
    \Lagr_{smooth} = \frac{1}{|\mathcal{G}|} \sum_{\textbf{x} \in \mathcal{G}} \Delta_x^2 + \Delta_y^2 + \Delta_z^2,
\end{align}
where  
    $\Delta_{x,y,z} = V_\alpha(\textbf{x} + \epsilon_{x,yz}) - V_\alpha(\textbf{x})$
is the feature difference of some sampled vertices.

\subsection{Reconstruction Uncertainty Learning}
\label{sec:method:uncertainty}
Recent studies \cite{zhan2022activermap, goli2023bayes, pan2022activenerf, ran2022neurar, yan2023active} have investigated various approaches for quantifying uncertainty in implicit representations.
\cite{pan2022activenerf, ran2022neurar} propose implicitly learning uncertainty through an MLP network.
This uncertainty MLP predicts point uncertainties for each sampled point along selected rays.
These point uncertainties are then integrated to calculate the photometric uncertainty of each pixel, employing the volume rendering technique described in \cref{sec:method:neural_3d_mapping}.
However, this form of uncertainty, as noted in \cite{klodt2018supervising}, does not strongly correlate with geometric uncertainty.
Alternatively, \cite{zhan2022activermap} opts for explicit and efficient computation of geometric uncertainty, represented as a 3D volume, from predicted densities.
Notably, the methods mentioned above are either RGB-based, lacking depth sensing, or do not incorporate depth measurements in uncertainty learning. 
This omission is significant, as depth information is essential for accurate uncertainty quantification.
In our work, we integrate the uncertainty learning process with depth rendering, as outlined in \cref{eqn:depth_loss}, within the bundle adjustment framework.
This integration follows the strategy proposed in \cite{kendall2017uncertainties}, effectively combining depth data with uncertainty.
\begin{align}
    & \Lagr_{d} = \frac{1}{|R_d|} \sum_{r \in R_d}
            \left (
            \frac{1}{2 \hat{\sigma}_r^2} (\hat{d}_r - D_r)^2 + \frac{1}{2} \text{log} \hat{\sigma}_r^2
            \right )
            \label{eqn:depth_uncert_loss}, \\
            & \text{where } 
            \hat{\sigma}_r^2 = \frac{1}{\sum_{i=1}^{M} w_i} \sum_{i=1}^{M} w_i \sigma^2_i
\end{align}

This study delves into two distinct methodologies for representing reconstruction uncertainty: implicit and explicit representations.
For the implicit approach, we employ an MLP to estimate point uncertainty, 
$f_{\sigma}(\gamma(\textbf{x}), \textbf{h}) \mapsto V_\sigma(\textbf{x})$.
However, our observations highlight a notable drawback of this implicit uncertainty representation.
Due to the reliance on the UncertaintyNet for predictions, any parameter update within the MLP results in alterations to uncertainty values across all regions, including those yet to be observed, 
\ie regions that lack observations are expected to exhibit high uncertainty; however, these areas often show random uncertainty levels instead.
In response to this challenge, we develop a learnable uncertainty volume, $V_\sigma$, designed to represent surface reconstruction uncertainty efficiently.
This volume enables rapid querying of uncertainties via trilinear interpolation,  $\sigma^2_i = f_\rho (V_\sigma (\textbf{x}_i))$, 
followed by a non-linear softplus activation function $f_\rho(.)$.
We initially set the volume with high uncertainty.
Significantly, as this volume is updated during bundle adjustment through uncertainty-aware depth rendering, only the uncertainties in regions that have been observed are modified.
This feature is vital for the effectiveness of active vision tasks.
The comparative advantages of our explicit representation over implicit methods are further detailed in \cref{sec:exp:ablation}.

\subsection{Uncertainty-aware Planning}
\label{sec:method:active_plan}

\begin{figure}[t!]
        \centering
  	\includegraphics[width=1.0\columnwidth]{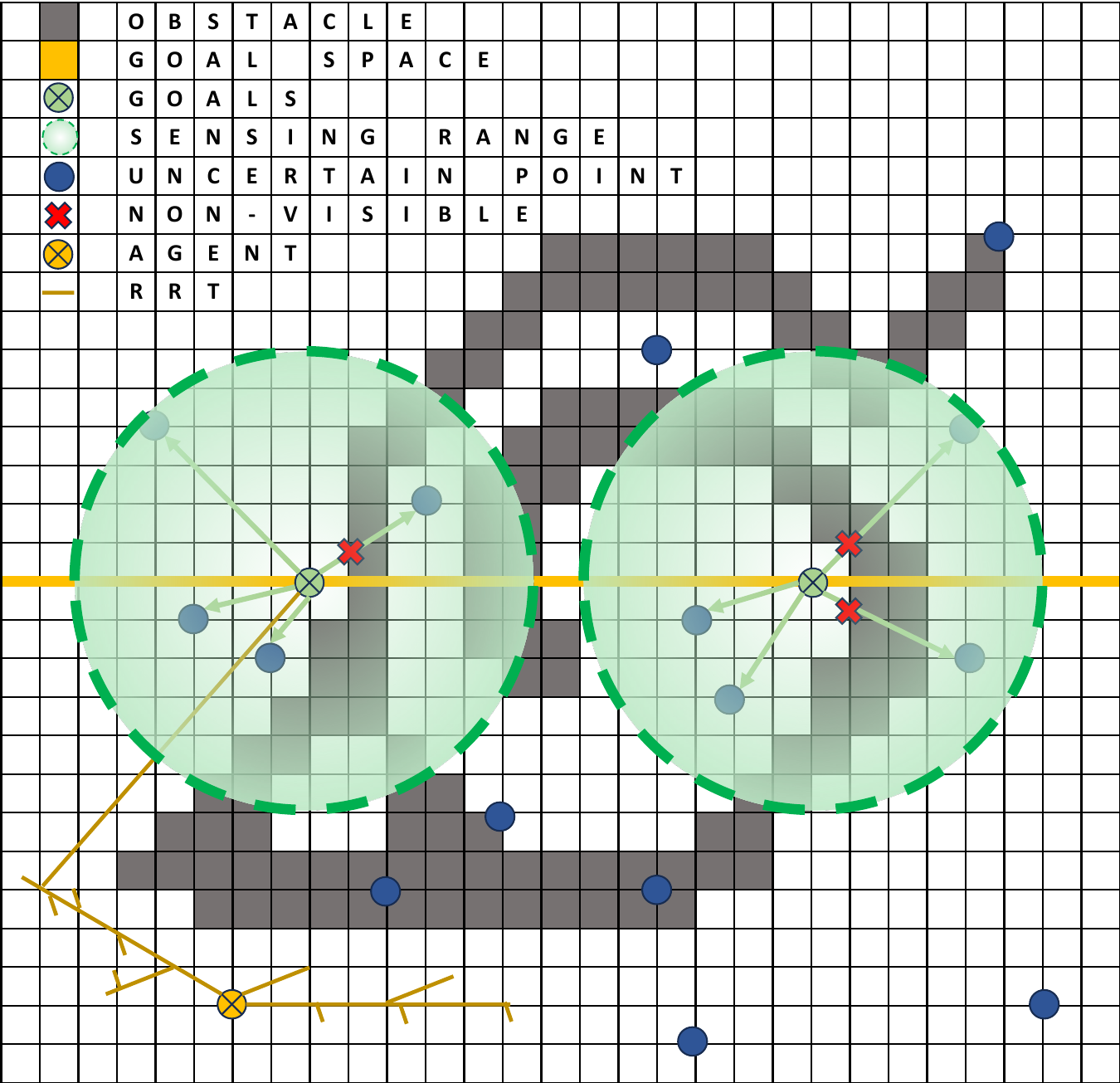}
		\caption{
            \textbf{Uncertainty-aware Planning Illustration}.
            The top-\textit{k} uncertain points are accumulated within the sensing range at each potential goal location. 
            The goal with the greatest level of uncertainty is subsequently selected as the provisional target location.
            Efficient RRT planning effectively identifies a viable trajectory from the agent's current position to the designated goal.
            }
		\label{fig:active_plan}
		\vspace{-10pt}
\end{figure}

In this section, we elaborate on the application of learned uncertainty and geometry in active planning, aiming to achieve comprehensive and high-quality reconstruction.
The planning module comprises two primary components: 
Goal Searching and Path Planning.
Utilizing the up-to-date SDF map that incorporates the learned geometric uncertainty, our primary goal is to pinpoint the most effective goal location for reducing overall map uncertainty.
To this end, we introduce an innovative uncertainty aggregation strategy, which facilitates the creation of an uncertainty-aware goal space.
Following the identification of the optimal observation location, we proceed with executing efficient path planning to establish a trajectory toward the chosen goal.
A 2D illustration of this approach is depicted in \cref{fig:active_plan}.

\paragraph{Uncertainty Aggregation for Goal Search}
Utilizing the most recent mapping model, denoted as $\textbf{M}$, we undertake two key constructions.
First, we generate an SDF volume, $V_s \in \mathbb{R}^{H \times W \times D}$, through uniform querying $\textbf{M}$ across the space.
Second, we establish an uncertainty volume, $V_\sigma \in \mathbb{R}^{H \times W \times D}$, which encapsulates the geometric uncertainty of the reconstruction space.
The foremost goal of this process is to determine the optimal observation location.
This location is characterized as the point from which the most substantial regions of high uncertainty can be observed.
To effectively identify such a location, we have developed a novel \textit{uncertainty aggregation} strategy.

Initially, we set up a multi-level Goal Space, denoted as $\textbf{S}_g \in \mathbb{R}^{H \times W \times N}$, comprising layers that are distributed at different heights within the space.
The arrangement is such that each layer is approximately 1 meter apart from its adjacent layers, providing a structured vertical distribution throughout the space.
Rather than aggregating uncertainties at every vertex within $V_\sigma$ onto the Goal Space, our method focuses on a set of vertices with the top-\textit{k} uncertainty, denoted as $\{\textbf{x}_\sigma\}^k$, where $k=300$.
For each point $\textbf{x}_g$ sampled within the Goal Space, we accumulate the uncertainty of all visible $\{\textbf{x}_\sigma\}^k$ points, provided they fall within the optimal observation range of $[0.5, 2]m$.
Visibility is ascertained by examining the SDF values between $\textbf{x}_g$ and $\textbf{x}_\sigma$.
Upon completing this aggregation process, the goal with the highest aggregated value is subsequently selected as the provisional target location.
The goal state $\textbf{s}_g$ is defined as the goal location looking at its most uncertain region.


\begin{algorithm}  [t]
    \caption{NARUTO: Neural Active Reconstruction}
  \begin{algorithmic}[1]
    \STATE \textbf{Initialization} 
        Mapping Model $\textbf{M}$ with [$V_s; V_\sigma]$; 
        Agent State $\textbf{s}_t=\textbf{s}_{0}$; 
        Goal Space $\textbf{S}_g$; 
        Observations $\{O\}_{i=0}^{0}$; 
        \textbf{PLAN\_REQUIRED} = True
        \FOR {$t \leftarrow \text{0 to T}$}
            \IF{\textbf{PLAN\_REQUIRED}} 
                \STATE \# Search a new goal from Goal Space if needed
                \STATE \textbf{GoalSearch}($\textbf{M}_t$, $\textbf{s}_t$) $\rightarrow$ $\textbf{s}_g \in \textbf{S}_g$ 
                
                \STATE \# Plan a feasible path based on $M_t$ towards $\textbf{s}_g$
                \STATE \textbf{PathPlanning}($\textbf{M}_t$, $\textbf{s}_t$, $\textbf{s}_g$) $\rightarrow$ $\{\textbf{s}_j\}_{j=t}^g$

                \STATE \# Set \textbf{PLAN\_REQUIRED} to False
                \STATE \textbf{PLAN\_REQUIRED} $\leftarrow$ False
            \ENDIF

            \STATE \# Execute to follow planned path
            \STATE \textbf{Action}  $\textbf{s}_t \leftarrow \{\textbf{s}_j\}_{j=t}^g$
            
            \STATE \# Update Database in keyframe steps
            \STATE \textbf{Observation}: acquire a new observation $O_t$
            \STATE \textbf{Update database}: $\{O\}_{i=0}^{t} \leftarrow \{O\}_{i=0}^{t-1}$

            \STATE \# Update Mapping Model
            \STATE \textbf{Mapping Optimization}: Update $\textbf{M}_{t} \leftarrow \textbf{M}_t$ 

            \STATE \# Replanning if detected collision or reached goal
            \STATE \textbf{CheckPlanRequired}: update \textbf{PLAN\_REQUIRED}
        \ENDFOR
  \end{algorithmic}
\label{alg:active_recon}
\end{algorithm} 
  

\paragraph{Efficient RRT Path Planning}
Upon pinpointing the goal location, our path planning module is activated to devise a viable path linking the current state, $\textbf{s}_t$, with the goal state, $\textbf{s}_g$.
For this purpose, we adopt a sampling-based path planning methodology akin to the Rapid-exploration Random Tree (RRT) \cite{lavalle2001rrt}, utilizing the SDF map $V_s$ as a basis.
Notably, executing the conventional RRT within a large-scale 3D environment proves to be considerably time-consuming.
To mitigate this challenge, we implement an efficient planning approach inspired by \cite{kuffner2000rrtconnect}.
Our strategy enhances the traditional RRT by not only iterating through random point sampling but also consistently seeking direct, feasible lines connecting these sampled points with the goal state.
Such augmentation significantly expedites the planning process, thereby making RRT practical and efficient even in expansive scenes.
Note that occasionally, the identified goal state $\textbf{s}_g$ may be situated in a location that, while lying within the predefined 3D bounding box, is actually outside the navigable space.
In such instances, RRT typically fails to find a valid or feasible path, as shown by reaching the maximum sampling number. 
To address this issue, we assess the reachability of all $V_\sigma$ vertices by querying RRT.
If a vertex is determined to be unreachable — specifically, if it lies at a minimum distance beyond the agent's step size — it is then excluded from the uncertainty aggregation process.

\paragraph{Action Execution}
In our system, the agent is capable of performing several actions under various events:
\begin{itemize}
\item \textbf{\textit{Move}}: The agent moves towards the target, looking at the 3D point with the highest uncertainty.
\item \textbf{\textit{Observe}}: Upon reaching $\textbf{s}_g$, the agent sequentially observes the top-\textit{10} uncertain points within the sensing range via rotational motion.
\item \textbf{\textit{Stay}}: The agent remains stationary either upon reaching the goal location or when collisions are detected.
\end{itemize}
Note that Goal Space and the RRT space can be tailored to suit the specific dimensions of the scene as well as the type of agent involved, whether it be a ground robot or an aerial robot. 
To demonstrate the generalization of our system, we model the agent as a free-moving entity with a spherical body, which has a radius of 5cm.
The agent's motion is constrained to translations $\leq 10$cm and rotations $\leq 10^\circ$.

\begin{figure*}[th!]
        \centering
		\includegraphics[width=1.0\textwidth]    {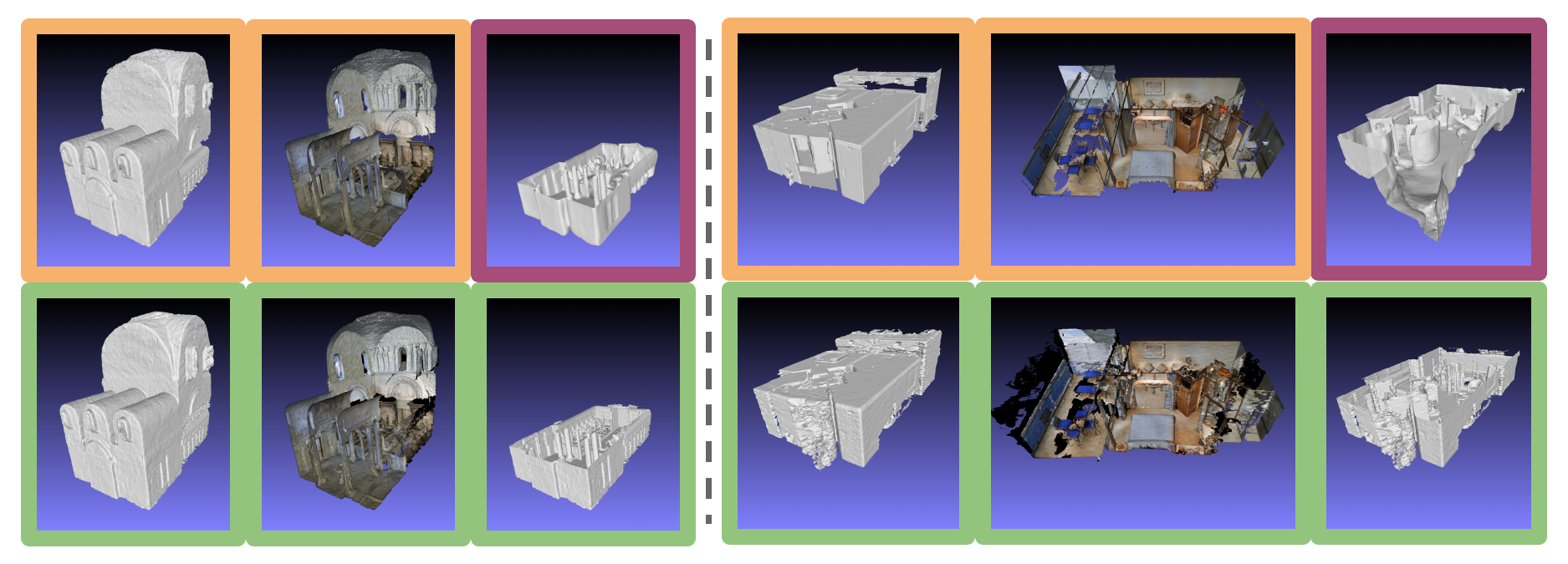}
		\caption{
            \textbf{
            Matterport3D Results} Two scenes (Left: pLe4; Right: HxpK) are presented here. 
            The results are distinguished by border colors: 
            [\textcolor[HTML]{F4B070}{Ground Truth }, \textcolor[HTML]{A24F78}{ANM}\cite{yan2023active}, \textcolor[HTML]{98C281}{Ours}].
            In our results, notably in the second and fifth columns, black regions signify incomplete GT mesh, illustrating the extrapolation capacity of our neural mapping module.
            Results in columns 3 and 6 are trimmed for better comparison.
            }
            
		\label{fig:mp3d_result}
		\vspace{-10pt}
\end{figure*}

\subsection{Active Ray Sampling}
\label{sec:method:active_ray}

In the process of mapping optimization, Co-SLAM \cite{wang2023coslam} employs a strategy of sampling $N$ rays from both the database and the most recent keyframe.
While this random sampling technique facilitates optimization across various regions, it occasionally leads to inconsistent results.
Moreover, this approach does not ensure that regions characterized by subpar reconstruction quality are adequately sampled.
By incorporating the learned uncertainty, we introduce a more targeted ray sampling method.
This approach retains the diversity of the original sampling strategy but enhances it by substituting $N'$ rays from the random sample with the top-$N'$ rays, selected based on their uncertainty.
This active ray sampling technique improves the consistency and quality of the system's output across different iterations, as presented in \cref{sec:exp:ablation}.

\subsection{Active Reconstruction}
\label{sec:method:act_recon}
Integrating the mapping module outlined in \cref{sec:method:neural_3d_mapping} and \cref{sec:method:uncertainty}, with the planning module from \cref{sec:method:active_plan}, we establish a comprehensive neural active reconstruction system, as detailed in \cref{alg:active_recon} and illustrated in \cref{fig:framework}.
Leveraging an up-to-date neural mapping model, this system employs the planning module to perform goal searching and path planning.
Subsequent to each action executed for acquiring a new RGB-D frame, a selection of rays from the keyframes is stored in a database to facilitate mapping optimization.
This storage occurs at a fixed interval of every \textit{5} steps.
Replanning is triggered under two conditions: either after the completion of the \textbf{\textit{Observe}} action at the goal location or upon detection of a collision.

\section{Experiments and Results}
\label{sec:exp}

\subsection{Experimental Setup}
\label{sec:exp:exp_setup}
\paragraph{Simulator and Dataset}

Our experiments utilize the Habitat simulator \cite{savva2019habitat} and are evaluated on two photorealistic datasets: Replica \cite{straub2019replica} and Matterport3D (MP3D) \cite{chang2017matterport3d}. 
Specifically, we select 8 scenes from Replica \cite{sucar2021imap} and 5 scenes from MP3D \cite{yan2023active} for our analysis. 
The experiments are designed to run for 2000 steps in Replica and 5000 steps in MP3D, reflecting the larger scene sizes in MP3D that necessitate more steps for thorough exploration.
In these experiments, our system processes posed RGB-D images at a resolution of $680 \times 1200$, with the field of view settings at $60^\circ$ vertically and $90^\circ$ horizontally.
We use 10cm as the voxel size for all experiments when generating 3D volume. 

This work represents a departure from previous neural active reconstruction efforts, which typically involve action spaces constrained to teleporting between discrete locations \cite{pan2022activenerf, ran2022neurar}, moving within limited areas such as a hemisphere \cite{zhan2022activermap}, or navigating the local vicinity on a 2D plane \cite{chaplot2020activeneuralslam, yan2023active}.
In contrast, we introduce the \textit{first} neural active reconstruction system operating with 6DoF movement in unrestricted 3D spaces.
Given the inherent randomness in the methods, we conduct each experiment five times to ensure reliability and present the average outcomes.
For experiments with active planning, the agent's starting position is randomly initialized within the traversable space for each trial.

\paragraph{Metrics}
We evaluate the reconstruction using 
\textit{Accuracy} (cm),
\textit{Completion} (cm),
\textit{Completion ratio} (\%) with a threshold of 5cm.
We also compute the mean absolute distance, \textit{MAD} (cm), between the estimated SDF distance on all vertices from the ground truth mesh. 
In line with methodologies employed in previous studies \cite{wang2022gosurf, wang2023coslam}, we refine the predicted mesh by removing unobserved regions and noisy points that are within the camera frustum but external to the target scene, utilizing a mesh culling technique.
Refer to \cite{wang2023coslam} for a detailed explanation of the mesh culling process.

\subsection{Evaluation}
\label{sec:exp:benchmark}

\begin{table}[t]
    \centering
    \resizebox{1\columnwidth}{!}{
    \begin{tabular}{l|c c c c}
        \hline
         & MAD (cm) $\downarrow$ & Acc. (cm) $\downarrow$ & Comp. (cm) $\downarrow$ & Comp. Ratio (\%) $\uparrow$ \\
        \hline
        FBE \cite{yamauchi1997frontier} &
        / & / & 9.78 & 71.18 \\
        UPEN \cite{georgakis2022uncertainty} &
        / & / & 10.60 & 69.06 \\
        OccAnt \cite{ramakrishnan2020occupancy} &
        / & / & 9.40 & 71.72 \\
        \textcolor[HTML]{A24F78}{ANM \cite{yan2023active}} &
        4.29 & 7.80 & 9.11 & 73.15 \\
        \textcolor[HTML]{98C281}{\textbf{Ours}} &
        \textbf{1.44} & \textbf{6.31} & \textbf{3.00} & \textbf{90.18} \\
        \hline
    \end{tabular}
    }
    \caption{ \textbf{MP3D Results}
    Our method shows superior performance with better reconstruction quality and completeness.
    }
    \label{tab:mp3d}
    \vspace{-10pt}
\end{table}

To our knowledge, this is the \textit{first} study to address the challenge of active surface reconstruction in large-scale indoor scenes with the provision for 6DoF movements in 3D space.
Previous studies that allow for 6DoF motions, such as \cite{kriegel2015efficient, isler2016information, lee2022uncertainty, ran2022neurar, zhan2022activermap}, have primarily focused on object-centric scenarios.
In contrast, earlier works targeting large-scale indoor scenes have generally been categorized under the \textit{active exploration} task.
These studies, including \cite{chaplot2020activeneuralslam, georgakis2022uncertainty, yan2023active}, often employ reinforcement learning-based planners and restrict agent movement to a 2D plane.
Notably, ANM \cite{yan2023active} is among the closest to our work; it also utilizes neural implicit representation for mapping in large-scale indoor environments.
Averaged results are presented in this section, while a comprehensive evaluation of individual scenes is included in the supplementary material. 

\paragraph{MP3D}
In \cref{tab:mp3d}, we provide a quantitative comparison of our system against previous studies on MP3D.
Our approach significantly surpasses prior work across all evaluation metrics.
The MAD metric reflects the precision of the learned 3D neural distance field in our model.
Furthermore, both the Completion and Completion Ratio metrics, which assess the extent of active exploration coverage in 3D space, indicate that our method achieves remarkably high completeness.
This success is attributable to our effective method of goal identification combined with the agent's unrestricted movement capabilities, as shown in \cref{fig:teaser}.

It is important to note that the Accuracy metric is calculated by computing the mean nearest distance (with respect to the prediction) between the predicted vertices and the ground-truth vertices.
However, a challenge arises with the MP3D scenes due to their real-world capture; the ground-truth mesh often exhibits incompleteness resulting from incomplete scanning.
In scenarios where neural implicit reconstruction is applied, the neural networks' extrapolation capacity can fill in these missing regions.
While this might be beneficial in some contexts, it poses a disadvantage for the Accuracy evaluation.
This effect is exemplified in \cref{fig:mp3d_result}, where the discrepancy due to neural network extrapolation is evident.
In \cref{fig:mp3d_result}, it is evident that our method yields a more comprehensive and high-fidelity reconstruction, underscoring the effectiveness of our approach.

\subsection{Ablation Studies}
\label{sec:exp:ablation}
\begin{table}[t]
    \centering
    \resizebox{1\columnwidth}{!}{
    \begin{tabular}{l| c c | c c | c c  }
        \hline
        \multirow{2}{*}{Method }
        & \multicolumn{2}{c|}{Acc. (cm) } 
        & \multicolumn{2}{c|}{Comp. (cm) } 
        & \multicolumn{2}{c}{Comp. Ratio (\%)} \\
         & $\mu$ & $\sigma^2 (10^{-3}$) 
         & $\mu$ & $\sigma^2 (10^{-3}$) 
         & $\mu$ & $\sigma^2 (10^{-2}$) 
         \\
        \hline
        \hline
        \multicolumn{7}{c}{\textbf{Neural SLAM}} \\
        \hline

        iMAP \cite{sucar2021imap}  &
        3.62 & / &
        4.93 & / &
        80.50 & / \\ 
        NICE-SLAM \cite{zhu2022niceslam}  &
        2.37 & / &
        2.63 & / &
        91.13 & / \\ 
        Co-SLAM \cite{wang2023coslam}  &
        \textbf{2.30} & 34.56 &
        \textbf{2.35} & 29.51 &
        \textbf{92.74} & 72.90 \\ 
        \cite{wang2023coslam}  \textbf{w/ ActRay} &
        \textbf{2.30} & \textbf{26.10} &
        \textbf{2.35} & \textbf{15.06} &
        92.70 & \textbf{11.77} \\

        \hline
        \hline
        \multicolumn{7}{c}{\textbf{Neural Mapping}: Tracking is disabled } \\
        \hline

        Co-SLAM \cite{wang2023coslam} 
        & \textbf{1.96} & 3.02
        & 2.00 & 0.86
        & 93.79 & 2.16 \\
        \cite{wang2023coslam}  \textbf{w/ ActRay}
        & \textbf{1.96} & \textbf{2.88}
        & \textbf{1.98} & \textbf{0.50}
        & \textbf{93.90} &\textbf{1.88} \\
        
        \hline
        \hline
        \multicolumn{7}{c}{\textbf{Neural Active Mapping}} \\
        \hline
        \textcolor[HTML]{D3A7BC}{w/o ActiveRay} &
        \multicolumn{2}{c|}{1.67} &
        \multicolumn{2}{c|}{1.76} &
        \multicolumn{2}{c}{96.89} \\
        \textcolor[HTML]{6BA2E8}{Uncertainty Net} &
        \multicolumn{2}{c|}{1.69} &
        \multicolumn{2}{c|}{2.05} &
        \multicolumn{2}{c}{94.62} \\
        \textcolor[HTML]{98C281}{\textbf{Full}} &
        \multicolumn{2}{c|}{\textbf{1.61}} &
        \multicolumn{2}{c|}{\textbf{1.66}} &
        \multicolumn{2}{c}{\textbf{97.20}} \\
        \hline

    \end{tabular}
    }
    \caption{Evaluation and Ablation Studies on Replica.}
    \label{tab:replica_ablation}
    \vspace{-5pt}
\end{table}

Replica features photorealistic 3D indoor scenes, spanning both room and building scales.
Each scene in this dataset is represented by a dense mesh, which typically exhibits greater completeness compared to the MP3D scenes.
Given this higher level of completeness, we primarily conduct our ablation studies on the Replica dataset to ensure more representative and robust results.

\paragraph{Active Ray Sampling}
\begin{figure}[t!]
        \centering
		\includegraphics[width=1.0\columnwidth]{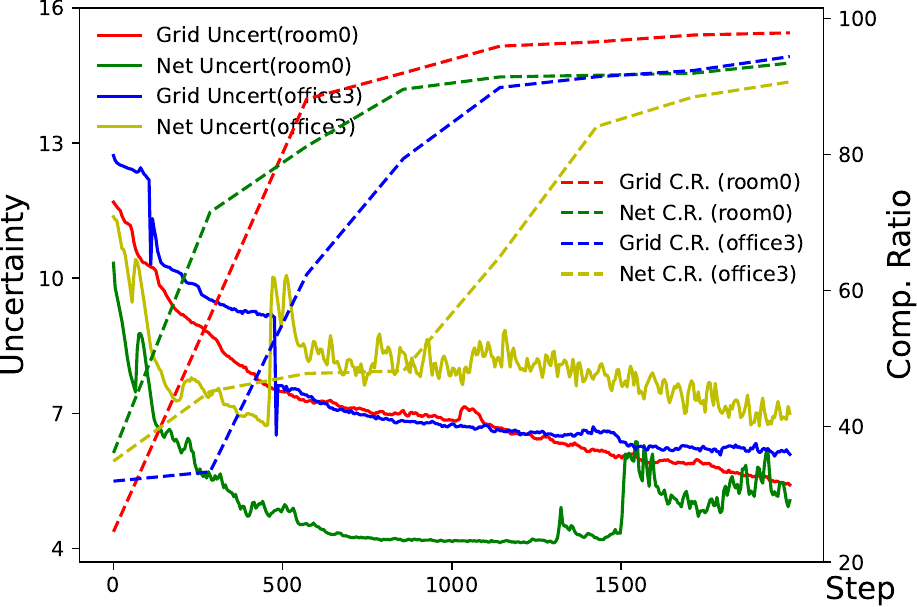}
		\caption{
                Evolution of Uncertainty and Completion Using Explicit Grid and Implicit Net. The abrupt decrease in \textit{Grid Uncert(office3)} correlates with the implementation of the reachability filtering strategy, as outlined in \cref{sec:method:active_plan}.
                }
		\label{fig:grid_vs_net}
		\vspace{-8pt}
\end{figure}
\begin{figure*}[t!]
        \centering
		\includegraphics[width=1.\textwidth]{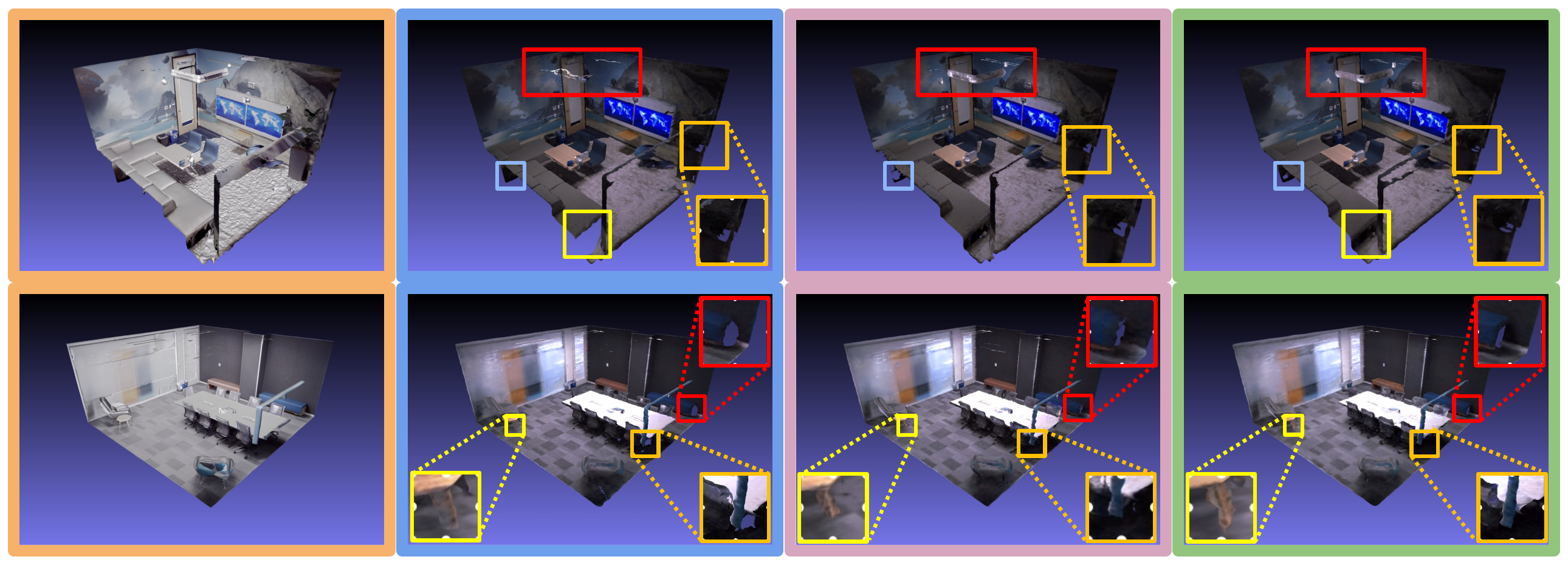}
		\caption{
            \textbf{Replica Results} 
            Two scenes (office0, office3) are shown in the first and second rows, respectively.
            The results represent 
            [
            \textcolor[HTML]{F4B070}{Ground Truth},
            \textcolor[HTML]{6BA2E8}{Uncertainty Net},
            \textcolor[HTML]{D3A7BC}{w/o ActiveRay},
            \textcolor[HTML]{98C281}{Full}
            ].
            Our \textcolor[HTML]{98C281}{Full} method shows a better completeness and quality on the 
            highlighted regions.
            Note that the GT visualization uses view-dependent rendering, unlike our mapping backbone, resulting in color differences in the visualizations.
            }
            \label{fig:replica_result}
		\vspace{-10pt}
\end{figure*}

In this section, we assess the efficacy of the Active Ray Sampling strategy (ActiveRay), as detailed in \cref{sec:method:active_ray}.
We tested the strategy across three distinct tasks, presenting the results in \cref{tab:replica_ablation}.
Leveraging our learned uncertainty, the Active Ray Sampling module acts as a versatile plug-and-play enhancement for existing neural mapping methods, leading to improved reconstruction outcomes.
Focusing on the Neural SLAM task, we integrate our learned uncertainty and the Active Ray Sampling strategy into Co-SLAM \cite{wang2023coslam}.
Our results demonstrate reconstruction quality comparable to the original Co-SLAM.
More importantly, multiple trials reveal that the inclusion of Active Ray Sampling yields more consistent results with reduced variance.
The Neural SLAM task, which involves estimating camera poses, introduces an additional complexity to the optimization process.
In the second task, we concentrate on mapping capabilities, deactivating the tracking function in Co-SLAM \cite{wang2023coslam}.
Without the instability introduced by the tracking thread, our method exhibits improved reconstruction quality compared to Co-SLAM.
A key advantage of this approach in both tasks is the enhancement of result stability, evidenced by reduced variance.
In the third task, focusing on Active Neural Mapping, we demonstrate that ActiveRay is a crucial element of our system.
We surmise that this effectiveness stems from our system's deliberate focus on accruing more observations from regions of uncertainty. 
Consequently, this leads to an increase in the number of valid rays, especially those marked by uncertainty, making them prime candidates for selection by ActiveRay.
We provide a qualitative comparison in \cref{fig:replica_result}, contrasting results obtained using our complete method with those achieved without ActiveRay.
The full implementation of our method, employing ActiveRay, demonstrates enhanced completeness and finer detail in thin structures.

\paragraph{Explicit Grid v.s. Implicit Net}
We discuss the use of explicit and implicit representation in \cref{sec:method:uncertainty}. 
It was noted that utilizing an implicit representation (Uncertainty Net) for learning uncertainty presents stability challenges. 
The optimization process employing Uncertainty Net is depicted in \cref{fig:grid_vs_net}, where it is juxtaposed with our proposed Uncertainty Grid for comparative analysis.
Two principal observations emerge from this comparison:
Firstly, both Uncertainty Net and Uncertainty Grid demonstrate rapid convergence, underscoring the efficacy of our uncertainty-aware planning approach. 
Secondly, as previously discussed in \cref{sec:method:uncertainty}, Uncertainty Net tends to produce fluctuating uncertainty values during the optimization phase due to continuous updates in network parameters. 
This instability is also illustrated in \cref{fig:grid_vs_net}, where we include 
$\text{log}(\sum_{\textbf{x}_i} V_\sigma (\textbf{x}_i))$ and the completion ratios, 
highlighting the comparative stability offered by Uncertainty Grid.
In Uncertainty Grid, a clear correlation is observed: the completion ratio increases as uncertainty decreases.
Conversely, in Uncertainty Net, these two metrics do not exhibit a strong correlation.
In \cref{fig:replica_result}, we present a qualitative comparison demonstrating that using Uncertainty Grid results in higher reconstruction completeness than Uncertainty Net.


\section{Discussion}
\label{sec:conclusion}

In summary, \titlePrefix{} represents a significant advancement in the field of neural active reconstruction. 
By integrating a hybrid neural representation with uncertainty learning, and a novel uncertainty-aware planning module, 
we present the \textit{first} neural active reconstruction system that enables agents to execute 6DoF movement in unrestricted space. 
Furthermore, the enhancement of state-of-the-art neural mapping methods through our active ray sampling strategy underscores the versatility and practicality of \titlePrefix{}. 
Rigorous evaluation in diverse environments using an indoor scene simulator demonstrates our system's superior performance, outperforming existing methods on benchmark datasets such as Replica and MP3D, setting a new standard in active reconstruction.

While \titlePrefix{} exhibits outstanding performance, future research directions are identified to advance the field. 
Firstly, the current assumption of known localization and perfect action execution, which might not hold in real-world scenarios, suggests the need for a robust planning and localization module to enhance real-world applicability. Secondly, the agent's motion constraints, vital in practical applications, should be considered to refine the system's general movement solution. Lastly, the use of a single-resolution uncertainty grid, primarily focusing on scene completeness, could be evolved into a multi-resolution uncertainty representation to meet diverse requirements. These future explorations aim to augment \titlePrefix{}'s practicality and adaptability in real-world settings, pushing the boundaries of autonomous robotic systems.

\newpage

\clearpage
\setcounter{page}{1}
\maketitlesupplementary

\section{Overview}
In this supplementary material, we provide a detailed outline structured as follows:
\cref{supp:sec:implementation} delves into additional implementation specifics of \titlePrefix{}.
\cref{supp:sec:runtime} examines the computation costs associated with each module.
Complementing the results in \cref{sec:exp},  \cref{supp:sec:more_dataset_results} extends our analysis with per-scene evaluations for MP3D and Replica.

\section{Implementation Details}
\label{supp:sec:implementation}

\paragraph{Hardware Details}
We run the experiments on a desktop PC with a 2.2GHz Intel Xeon E5-2698 CPU and NVIDIA V100 GPU.

\paragraph{Memory requirement}
Memory consumption varies depending on the scene size. As a reference, in a $120 m^3$ scene, the corresponding GPU memory and RAM are 8.1GB and 8.6GB respectively. 
The consumption can be further reduced with a more efficient implementation as our current implementation involves intensive exchanges between RAM and GPU memories.

\subsection{Neural Mapping Details}
We adopt Co-SLAM~\cite{wang2023coslam} as the foundational mapping framework for our system, adhering to the hyperparameter configurations established therein. 
For details pertaining to the hyperparameters specific to the mapping component, we direct readers to \cite{wang2023coslam} for comprehensive information.

\subsection{Efficient RRT Details} \label{sec:supp:efficient_rrt}
Path planning in three-dimensional spaces presents significant computational challenges, particularly when employing standard 3D RRT algorithms \cite{lavalle2001rrt}. 
In our approach, we introduce an accelerated version of RRT, dubbed E-RRT (Efficient RRT), which incorporates several optimizations for improved performance.

The primary innovation in E-RRT, drawing inspiration from RRT-Connect~\cite{kuffner2000rrtconnect}, is its strategy to first attempt direct connections from the growing tree to the goal at each iteration. 
While this does not ensure the shortest path, it significantly enhances the efficiency of finding a viable path.

Furthermore, E-RRT enhances the process of node expansion. 
Instead of adding a single node, our method integrates a series of feasible points uniformly distributed between a randomly generated node and its nearest neighbor in the tree, based on a predefined step size, for instance, 10 cm, up the distance of $M \times \text{step size}$. Here $M$ equals to 10.
This modification substantially accelerates the expansion of the tree, especially in the initial growth stages.

Lastly, we address the increasing computational load associated with nearest-neighbor searches as the tree expands. 
By leveraging parallel processing on a GPU, E-RRT achieves a consistently high search speed, thus mitigating the computational costs that typically escalate with tree complexity.

\subsection{Collision Detection}
\begin{figure}[t!]
        \centering
		\includegraphics[width=1.0\columnwidth]{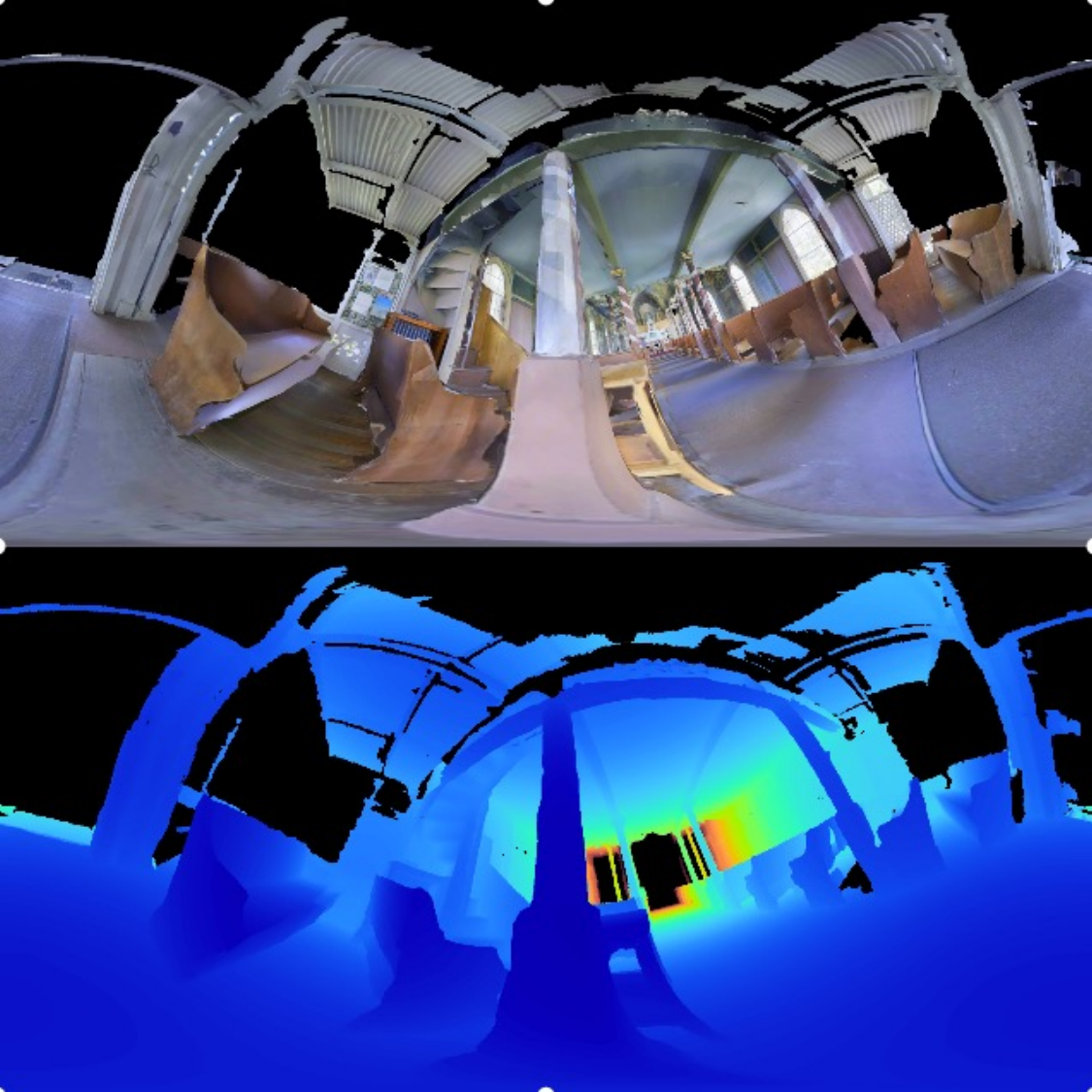}
		\caption{
                \textbf{Equirectangular RGB-D Example}
                Black regions refer to the invalid regions with zero depth measurement. 
                The ratio of black regions increases significantly when the agent leaves the building. 
                This is used as a signal for collision detection.
                }
		\label{fig:erp_eg}
\end{figure}
We have tailored two distinct collision detection methodologies to align with the nuances of the Replica and Matterport3D datasets. 

For experiments conducted within the Replica dataset, collision detection is facilitated through an SDF map derived from our hybrid scene representation. 
We assess potential collisions by sampling points at 2 cm intervals between consecutive states and querying the SDF map at these points. 
A collision is inferred when the SDF value at any point falls below the 5 cm threshold, consistent with our model of the agent as a sphere with a 5 cm radius.

This protocol effectively prevents the agent from intersecting with wall surfaces during simulations. Nonetheless, it cannot preclude the agent from exiting the scene through non-watertight boundaries. 
In contrast, the Matterport3D dataset, reflecting real-world environments, presents unique challenges with regions devoid of geometry—artifacts of incomplete depth data during dataset construction. 
These gaps in the environment can erroneously permit the agent to traverse through ``walls" or exit buildings. 
To counteract this, in addition to the SDF-based collision detection, we have developed a specialized collision detection system that assesses equirectangular depth measurements (\eg \cref{fig:erp_eg}) at prospective states, calculating the proportion of invalid regions. 
An increase in this proportion signals potential egress from the building, and by establishing a threshold ratio, we can determine the validity of the next state, thereby preventing unintended departure from the environment.

\subsection{Rotation Planning}
As delineated in \cref{sec:method:active_plan}, when the agent arrives at a designated goal state $\textbf{s}_g$, it proceeds to sequentially observe the top-10 points of uncertainty within its sensing radius through a series of rotational movements. 
In an effort to reduce the number of steps necessary to cover all ten of these uncertain perspectives, we have devised a straightforward rotational planning algorithm. 
This method involves identifying the subsequent viewpoint that can be reached with the least rotational effort and then executing the transition using a Spherical Linear Interpolation (SLERP) strategy.

\subsection{Active Ray Sampling Details}
In the context of mapping optimization within Co-SLAM\cite{wang2023coslam}, the conventional approach entails the random selection of 2048 pixels from the database, supplemented by a minimum of 100 pixels from the current viewpoint. 
Our Active Ray Sampling strategy introduces a refinement to this process. 
Specifically, we quadruple the count of randomly sampled pixels, thus drawing 8192 pixels from the database and ensuring at least 400 pixels from the current viewpoint. 
Within this augmented sample set, we then identify and prioritize the 500 most uncertain pixels. 
The remaining 1548 pixels are selected from the database, in addition to a minimum of 100 random points from the current viewpoint. 
This hybrid sampling method effectively combines the breadth of random sampling with the targeted insight of Active Ray Sampling, thereby capturing a broad yet informative snapshot of the environment.

\section{Runtime Analysis}
\label{supp:sec:runtime}

\begin{figure*}[th!]
        \centering
		\includegraphics[width=1.0\textwidth]{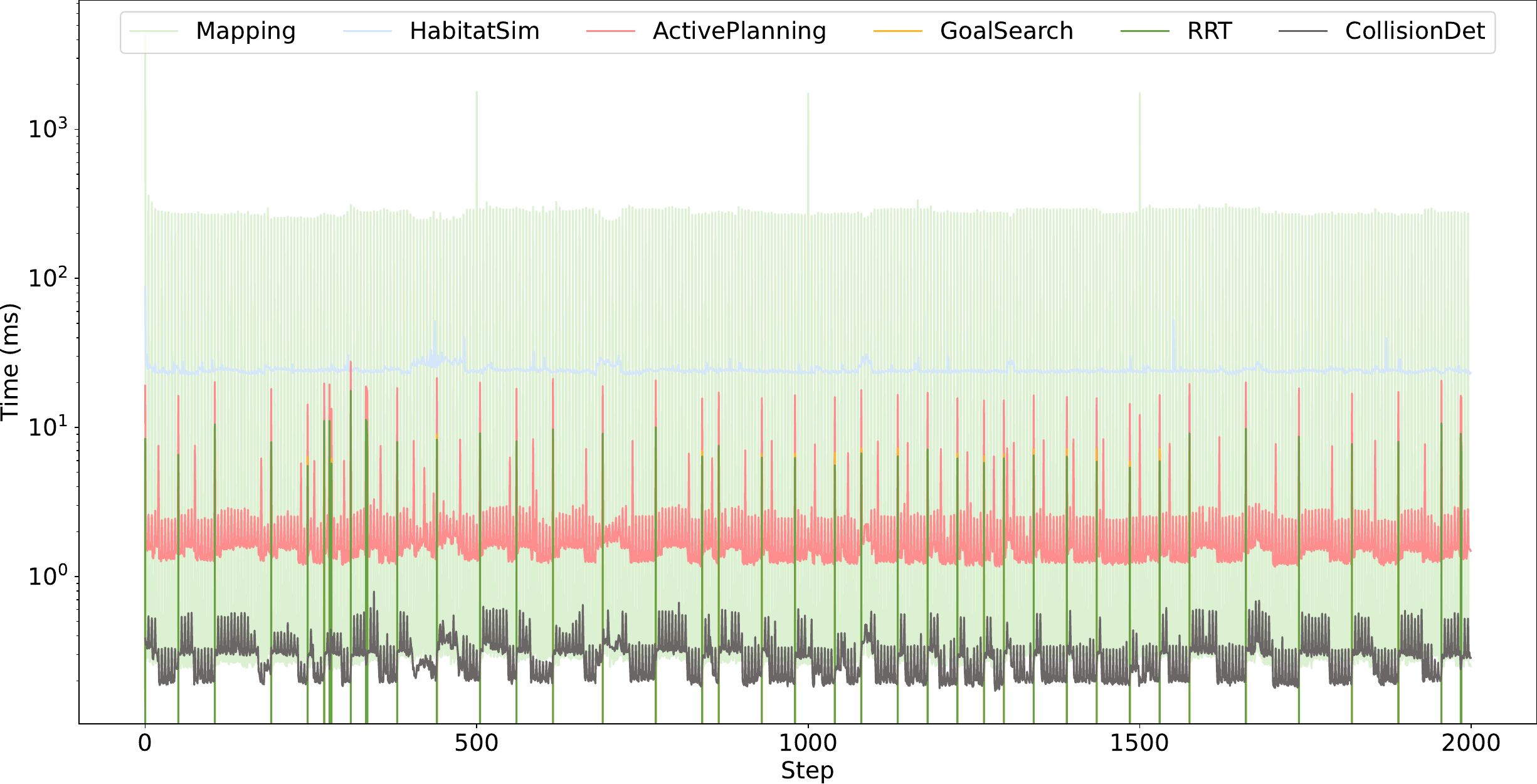}
		\caption{
            \textbf{Runtime Analysis in the Replica-room0 Environment}
            This figure illustrates the runtime analysis of each module within the \textit{Replica-room0} environment.
            A notable runtime impulse is observed during goal-searching iterations.
            The analysis encompasses three principal modules:  
            \textcolor[HTML]{D4E7FA}{Habitat Simulator} for data generation, 
            \textcolor[HTML]{FE8D8E}{Active Planning} for path planning, 
            and \textcolor[HTML]{DBF1D1}{Mapping} for mapping optimization.
            In the \textcolor[HTML]{FE8D8E}{Active Planning} module, further runtime analysis includes its submodules: \textcolor[HTML]{FEB62B}{Uncertainty-aware Goal Searching}, \textcolor[HTML]{67A246}{RRT Path Planning}, and \textcolor[HTML]{6A6666}{Collision Detection}.
            }
		\label{fig:runtime}
\end{figure*}

\subsection{System Runtime}
In this section, we present a detailed runtime analysis of the three major modules in \titlePrefix{}, as illustrated in \cref{fig:runtime}. 
The first module is a simulator for data generation. 
The second is a mapping module optimized for a hybrid scene representation. 
Lastly, we have an uncertainty-aware planning module. 
For data generation, HabitatSim requires, on average, 24.4ms to generate $680 \times 1200$ RGB-D data per iteration. 
The Mapping module, although taking about 300ms per iteration, averages 60.5ms since it is activated only every five keyframes. 
The Active Planning module averages 2.1ms, which includes 0.3ms for collision detection per iteration. 
Additionally, Active Planning encompasses two modules that are triggered occasionally when the 'PLAN\_REQUIRED' condition is met. 
These are the uncertainty-aware goal searching, averaging 6.8ms, and RRT path planning, averaging 5.77 ms. 
In conclusion, our analysis demonstrates that \titlePrefix{} offers real-time capabilities, particularly due to its efficient planning module.

\subsection{RRT Runtime Analysis}

\begin{table}[t]
    \centering
    \resizebox{1\columnwidth}{!}{
    \begin{tabular}{l|c c c}
        \hline
         Method & Time (ms) & Node Num. & Step Num.  \\
        \hline
        RRT & 
            $19 \times 10^{3} $ & $19 \times 10^{3}$ & $28 \times 10^{3}$ \\
        \hline
        w/o direct line & 
            $17 \times 10^{3} $ & $20 \times 10^{3}$ & $21 \times 10^{3}$ \\
        w/o fast tree & 
            $16.00$ & $44.17$ & $2.56$ \\
        Ours (E-RRT) & 
            $5.77$ & $16.70$ & $1.19$  \\
        
        \hline
    \end{tabular}
    }
    \caption{ \textbf{RRT runtime analysis on Replica-room0.}
    We conducted a runtime analysis of RRT variants, revealing that our optimized RRT implementation significantly outpaces traditional RRT in planning speed, achieving real-time planning capabilities.
    }
    \label{tab:rrt_runtime}
\end{table}

In this section, we delve deeper into our optimized RRT implementation, as outlined in \cref{sec:supp:efficient_rrt}. 
We have engineered a customized version of RRT that enhances planning speed through several strategies:
\begin{itemize}
    \item Direct Line: Actively identifying straight paths that link the RRT tree to the goal.
    \item Fast Tree: Speeding up the expansion of the tree.
    \item Parallel Computing: Utilizing GPU processing for increased efficiency.
\end{itemize}

These innovations significantly reduce the time required for path planning, making our RRT variant highly suitable for real-time applications. 
We present an ablation study on the runtime performance of our RRT approach in \cref{tab:rrt_runtime}. 
To maintain consistency, all experiments were conducted using parallel processing for nearest-neighbor searches during tree expansion.

\paragraph{Evaluation}
Our evaluation of the methods encompasses three key metrics: 
the average time taken for each path planning request, 
the average number of nodes generated within the RRT tree, 
and the average number of steps taken in the RRT process.

\paragraph{Analysis}
Compared to traditional RRT, our efficient RRT implementation is markedly faster, both in average planning time and iteration count. 
It also generates fewer nodes and uses less memory, as shown by the reduced average number of nodes required per planning request. 
The ablation study detailed in \cref{tab:rrt_runtime} highlights that our primary strategy for improvement involves identifying potential straight paths, drawing inspiration from RRT-Connect \cite{kuffner2000rrtconnect}. 
This approach, along with quicker tree growth, not only accelerates the planning process but also decreases memory usage.

\section{Additional Experimental Results}
\label{supp:sec:more_dataset_results}

\subsection{Detailed results on MP3D and Replica}
\label{supp:sec:detailed_results}
\begin{table*}[htbp]
    \centering
    \begin{tabular}{c|l*{8}{c}|c}
        \hline
        Method & Metrics & 
        office0 & office1 & office2 & office3 & office4 & room0 & room1 & room2 & 
        Avg. \\
        \hline
        \hline

        \multicolumn{11}{c}{
        \textbf{Neural SLAM}
        } \\
        \hline

        \multirow{3}{*}{
        Co-SLAM \cite{wang2023coslam}
        } 
        
        & Acc. [cm] $\downarrow$ & 
        1.68 & 1.46 & 2.98 & 3.07 & 2.44 & 2.14 & 2.64 & 2.02 & \textbf{2.30} \\
        & Comp. [cm] $\downarrow$ & 
        1.68 & 1.82 & 2.70 & 2.83 & 2.64 & 2.25 & 2.84 & 2.02 & \textbf{2.35} \\
        & Comp. Ratio $\uparrow$ & 
        96.25 & 94.44 & 89.80 & 90.82 & 91.59 & 94.61 & 90.32 & 94.09 & \textbf{92.74} \\
        \hline

        \multirow{3}{*}{\cite{wang2023coslam} \textbf{w/ ActRay}} 
        & Acc. (cm) $\downarrow$ & 
        1.61 & 1.48 & 2.96 & 3.12 & 2.43 & 2.17 & 2.58 & 2.00 & \textbf{2.30} \\
        & Comp. (cm) $\downarrow$ & 
        1.61 & 1.85 & 2.67 & 2.96 & 2.67 & 2.26 & 2.78 & 2.03 & \textbf{2.35} \\
        & Comp. Ratio $\uparrow$ & 
        96.24 & 94.44 & 90.61 & 89.85 & 91.51 & 94.66 & 90.23 & 94.08 & 92.70 \\
        \hline
        \hline

        \multicolumn{11}{c}{
        \textbf{Neural Mapping}: Tracking is disabled.
        } \\
        \hline

        \multirow{3}{*}{
        Co-SLAM \cite{wang2023coslam}
        } 
        
        & Acc. [cm] $\downarrow$ & 
        1.50 & 1.28 & 2.56 & 2.69 & 2.25 & 2.01 & 1.55 & 1.87 & \textbf{1.96} \\
        & Comp. [cm] $\downarrow$ & 
        1.48 & 1.61 & 2.17 & 2.52 & 2.47 & 2.13 & 1.71 & 1.88 & 2.00 \\
        & Comp. Ratio $\uparrow$ & 
        96.33 & 94.65 & 92.47 & 91.43 & 91.34 & 94.67 & 95.45 & 93.95 & 93.79 \\
        \hline

        \multirow{3}{*}{\cite{wang2023coslam} \textbf{w/ ActRay}} 
        & Acc. (cm) $\downarrow$ & 
        1.47 & 1.27 & 2.55 & 2.71 & 2.26 & 2.02 & 1.57 & 1.87 & \textbf{1.96} \\
        & Comp. (cm) $\downarrow$ & 
        1.47 & 1.59 & 2.13 & 2.55 & 2.49 & 2.07 & 1.71 & 1.85 & \textbf{1.98} \\
        & Comp. Ratio $\uparrow$ & 
        96.44 & 94.80 & 92.90 & 91.32 & 91.32 & 94.92 & 95.40 & 94.12 & \textbf{93.90} \\
        \hline
        \hline


        \multicolumn{11}{c}{
        \textbf{Neural Active Mapping}
        } \\
        \hline

        \multirow{3}{*}{
        \textcolor[HTML]{D3A7BC}{w/o ActiveRay}
        } 
        & Acc. (cm) $\downarrow$ & 
        1.29 & 1.05 & 2.17 & 2.86 & 1.72 & 1.56 & 1.24 & 1.46 & 1.67 \\
        & Comp. (cm) $\downarrow$ & 
        1.40 & 1.50 & 1.66 & 3.14 & 1.76 & 1.67 & 1.45 & 1.47 & 1.76 \\
        & Comp. Ratio $\uparrow$ & 
        97.92 & 95.87 & 98.04 & 90.68 & 98.09 & 98.31 & 97.62 & 98.55 & 96.89 \\
        \hline

        \multirow{3}{*}{
        \textcolor[HTML]{6BA2E8}{Uncertainty Net}
        } 
        & Acc. (cm) $\downarrow$ & 
        1.32 & 1.05 & 2.04 & 3.13 & 1.70 & 1.58 & 1.26 & 1.45 & 1.69 \\
        & Comp. (cm) $\downarrow$ & 
        2.12 & 2.01 & 2.73 & 2.50 & 2.07 & 1.90 & 1.58 & 1.56 & 2.06 \\
        & Comp. Ratio $\uparrow$ & 
        94.21 & 93.22 & 92.62 & 92.12 & 94.24 & 96.36 & 96.65 & 97.54 & 94.62 \\
        \hline

        \multirow{3}{*}{
        \textcolor[HTML]{98C281}{\textbf{Full}}
        } 
        & Acc. (cm) $\downarrow$ & 
        1.30 & 1.03 & 2.25 & 2.29 & 1.75 & 1.56 & 1.25 & 1.47 & \textbf{1.61} \\
        & Comp. (cm) $\downarrow$ & 
        1.39 & 1.53 & 1.69 & 2.27 & 1.79 & 1.68 & 1.43 & 1.48 & \textbf{1.66} \\
        & Comp. Ratio $\uparrow$ & 
        98.17 & 95.26 & 97.54 & 93.91 & 97.93 & 98.28 & 98.04 & 98.47 & \textbf{97.20} \\
        \hline

    \end{tabular}
    \caption{Per-scene quantitative results on Replica\cite{straub2019replica} dataset}
    \label{tab:replica_full}
\end{table*}

\begin{table}[t]
    \centering
    \resizebox{1\columnwidth}{!}{
    \begin{tabular}{c| c | *{5}{c}|c}
        \hline
        Method & Metric & 
        Gdvg & gZ6f & HxpK & pLe4 & YmJk & 
        \textbf{Avg.} \\
        \hline
        \multirow{4}{*}{ANM \cite{yan2023active}} 
        & MAD (cm) $\downarrow$ & 
        3.77 & 3.18 & 7.03 & 3.25 & 4.22 & 4.29  \\
        & Acc. (cm) $\downarrow$ & 
        5.09 & 4.15 & 15.60 & 5.56 & 8.61 & 7.80 \\
        & Comp. (cm) $\downarrow$ & 
        5.69 & 7.43 & 15.96 & 8.03 & 8.46 & 9.11   \\
        & Comp. Ratio $\uparrow$ & 
        80.99 & 80.68 & 48.34 & 76.41 & 79.35 & 73.15   \\
        \hline

        \multirow{3}{*}{Ours} 
        & MAD (cm) $\downarrow$ & 
        1.60 & 1.23 & 1.53 & 1.37 & 1.45 & \textbf{1.44} \\
        & Acc. (cm) $\downarrow$ & 
        3.78 & 3.36 & 9.24 & 5.15 & 10.04 & \textbf{6.31} \\
        & Comp. (cm) $\downarrow$ & 
        2.91 & 2.31 & 2.67 & 3.24 & 3.86 & \textbf{3.00} \\
        & Comp. Ratio $\uparrow$ & 
        91.15 & 95.63 & 91.62 & 87.76 & 84.74 & \textbf{90.18} \\
        \hline
    \end{tabular}
    }
    \caption{
    \textbf{Per-scene quantitative results on Matterport3D \cite{chang2017matterport3d} dataset}. 
    Our method achieves consistently better reconstruction than the state-of-the-art method ANM \cite{yan2023active}.
    }
    \label{tab:mp3d_full}
\end{table}

\begin{figure*}[th!]
        \centering
		\includegraphics[width=0.95\textwidth, height=0.95\textheight]    {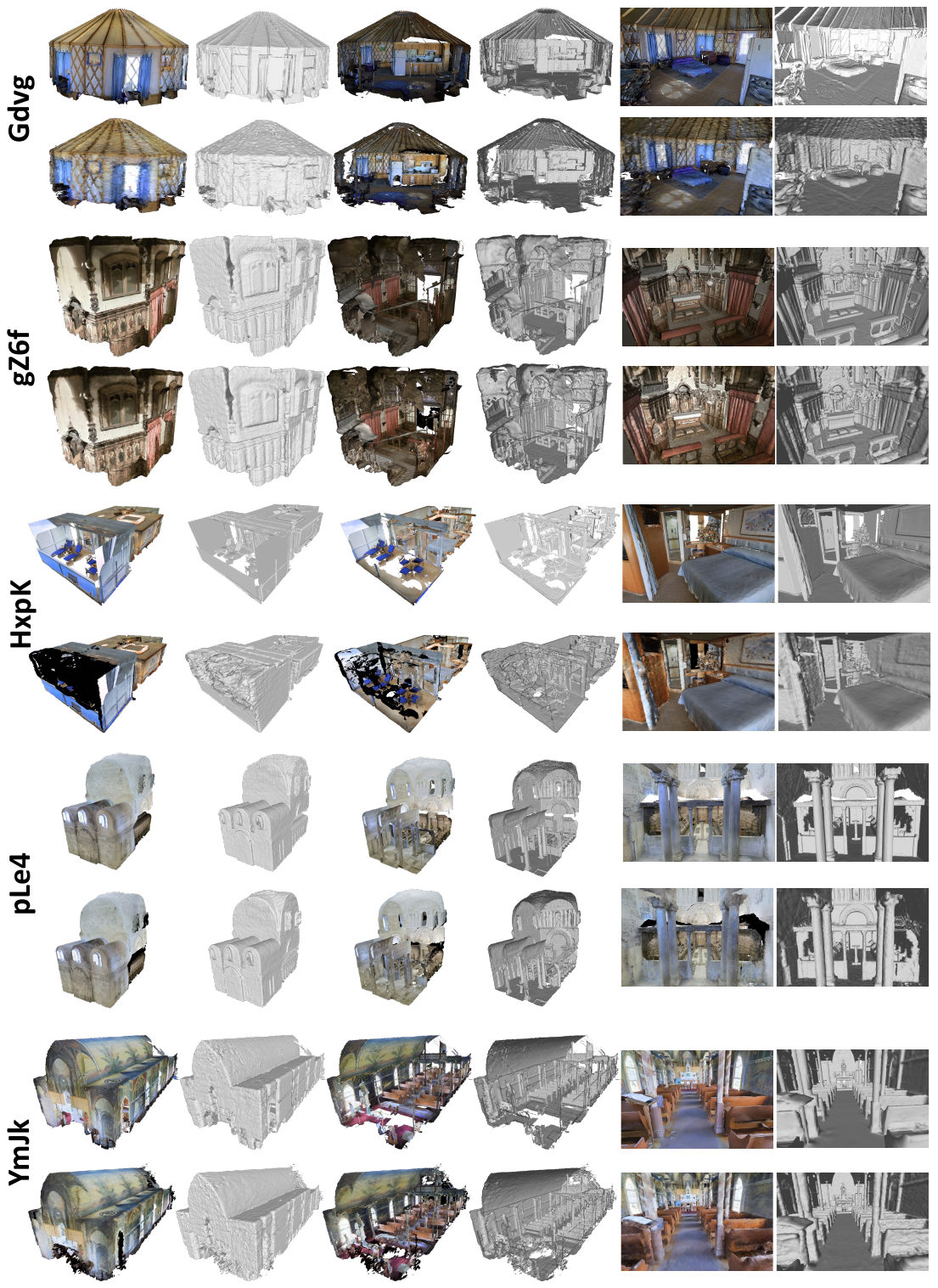}
		\caption{
\textbf{MP3D Reconstruction Results} 
 This presents a side-by-side comparison of the reconstruction results with the Matterport3D dataset. 
 The odd-numbered rows display the ground truth meshes, while the even-numbered rows feature the meshes reconstructed by our method. 
 Our results show a high level of quality and completeness, closely mirroring the ground truths. 
 This alignment underscores the efficacy of our method in accurately exploring and reconstructing complex spatial geometries.
            }
		\label{fig:mp3d_result_full}
\end{figure*}
\begin{figure*}[th!]
        \centering
		\includegraphics[width=0.95\textwidth, height=0.95\textheight]    {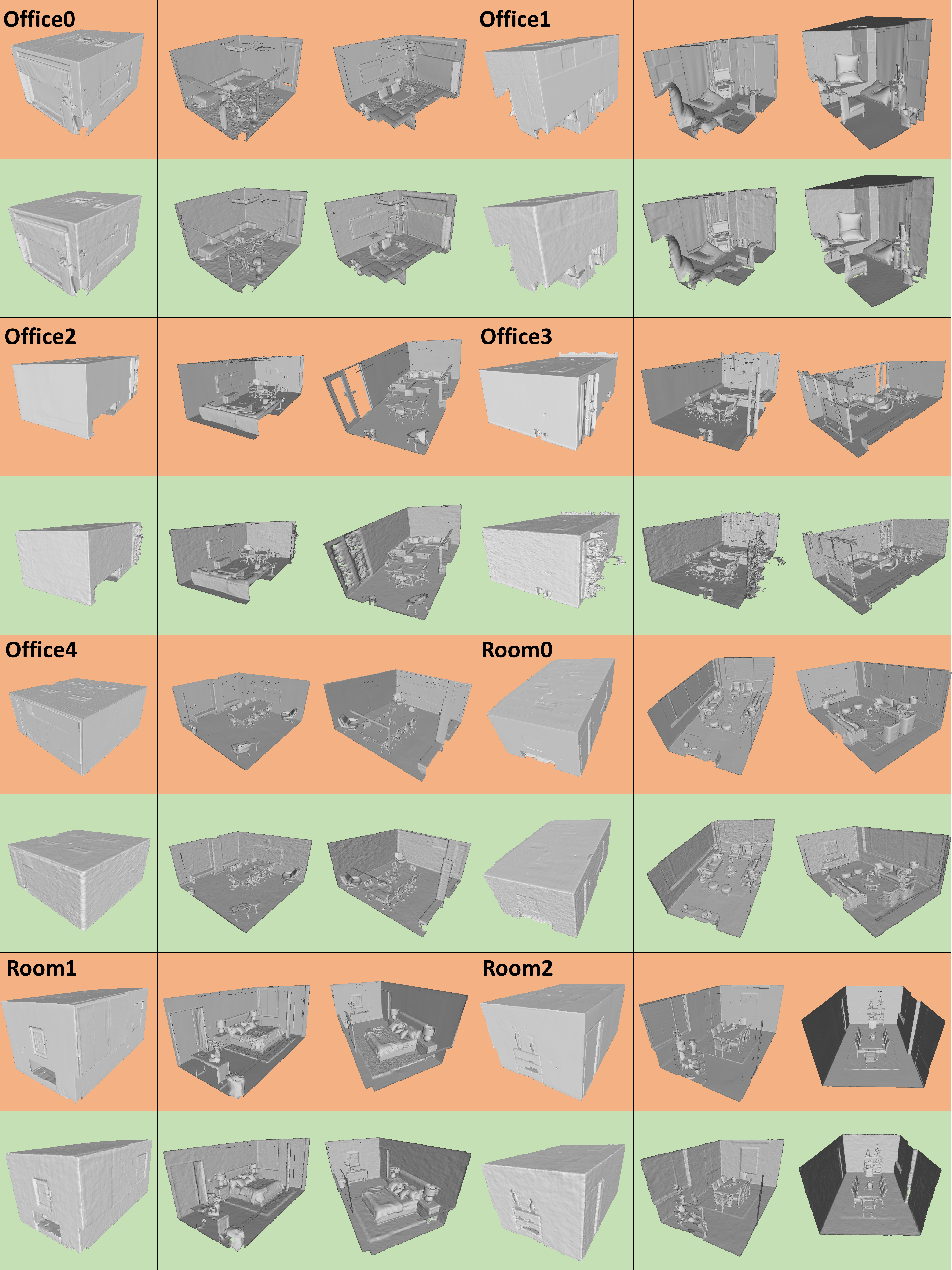}
		\caption{
\textbf{Replica Reconstruction Results} 
 This presents a side-by-side comparison of the reconstruction results with the Replica dataset. 
 The \textcolor[HTML]{F2AF86}{odd-numbered rows} display the ground truth meshes, while the \textcolor[HTML]{C8DFB6}{even-numbered rows} feature the meshes reconstructed by our method. 
 Our results show a high level of quality and completeness, closely mirroring the ground truths. 
            }
		\label{fig:replica_result_full}
\end{figure*}
\begin{figure*}[t!]
        \centering
		\includegraphics[width=1.\textwidth]{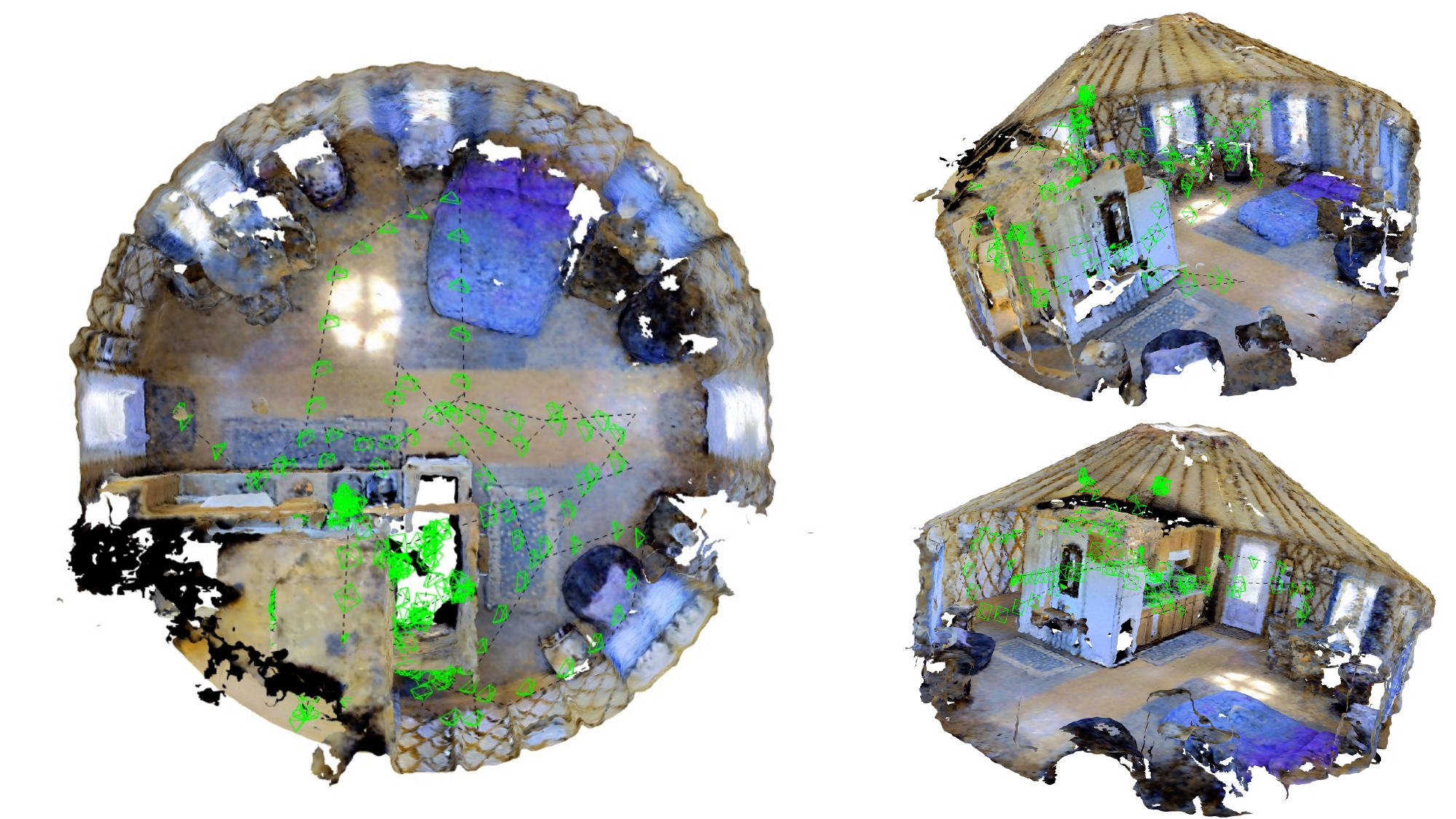}
		\caption{
            \textbf{Matterport3D (Gdvg)}
            Reconstructed Mesh and planned trajectory.
            }
            \label{fig:mp3d_Gdvg}
\end{figure*}

\begin{figure*}[t!]
        \centering
		\includegraphics[width=1.\textwidth]{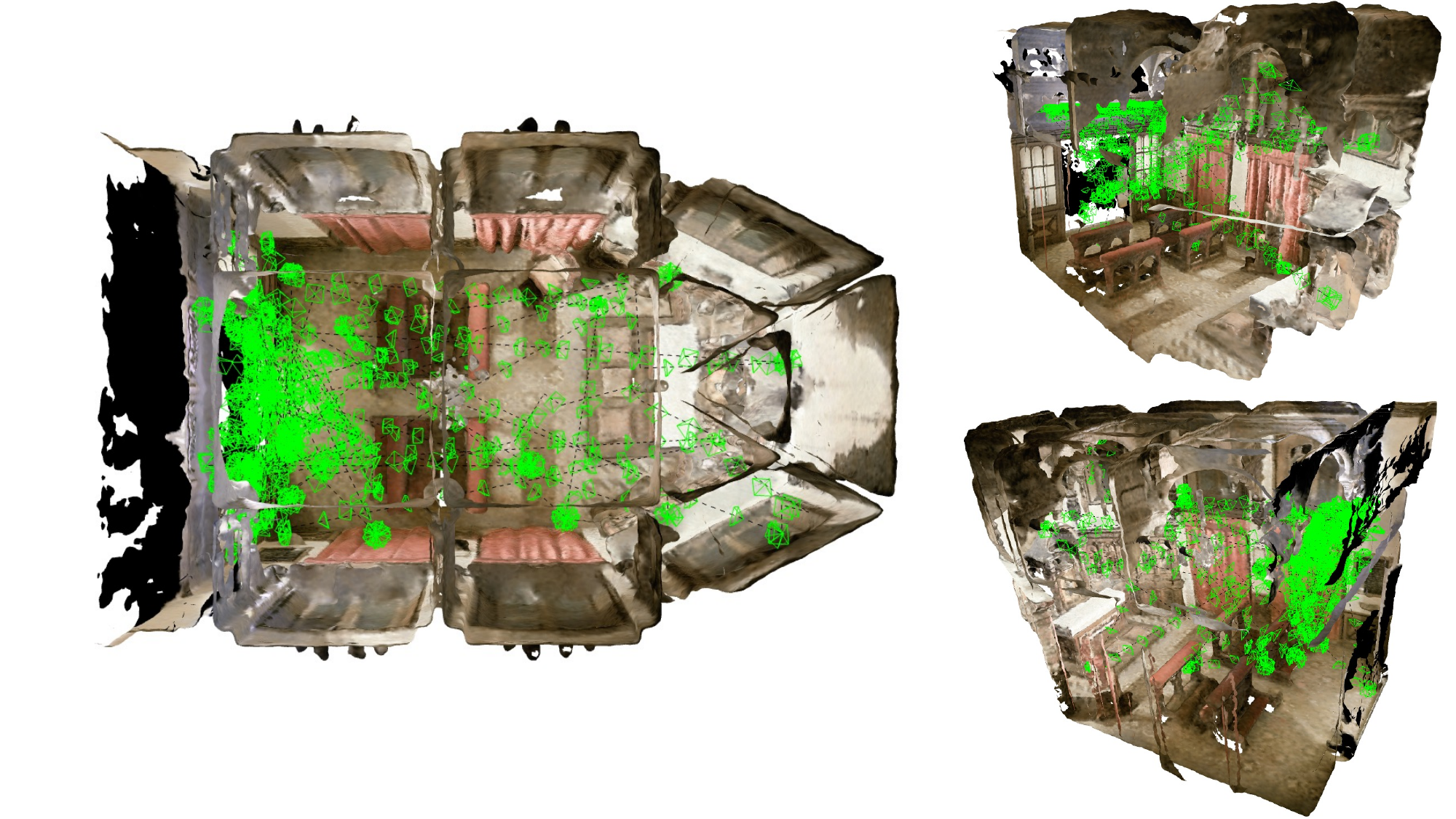}
		\caption{
            \textbf{Matterport3D (gZ6f)} 
            Reconstructed Mesh and planned trajectory.
            }
            \label{fig:mp3d_gZ6f}
\end{figure*}

\begin{figure*}[t!]
        \centering
		\includegraphics[width=1.\textwidth]{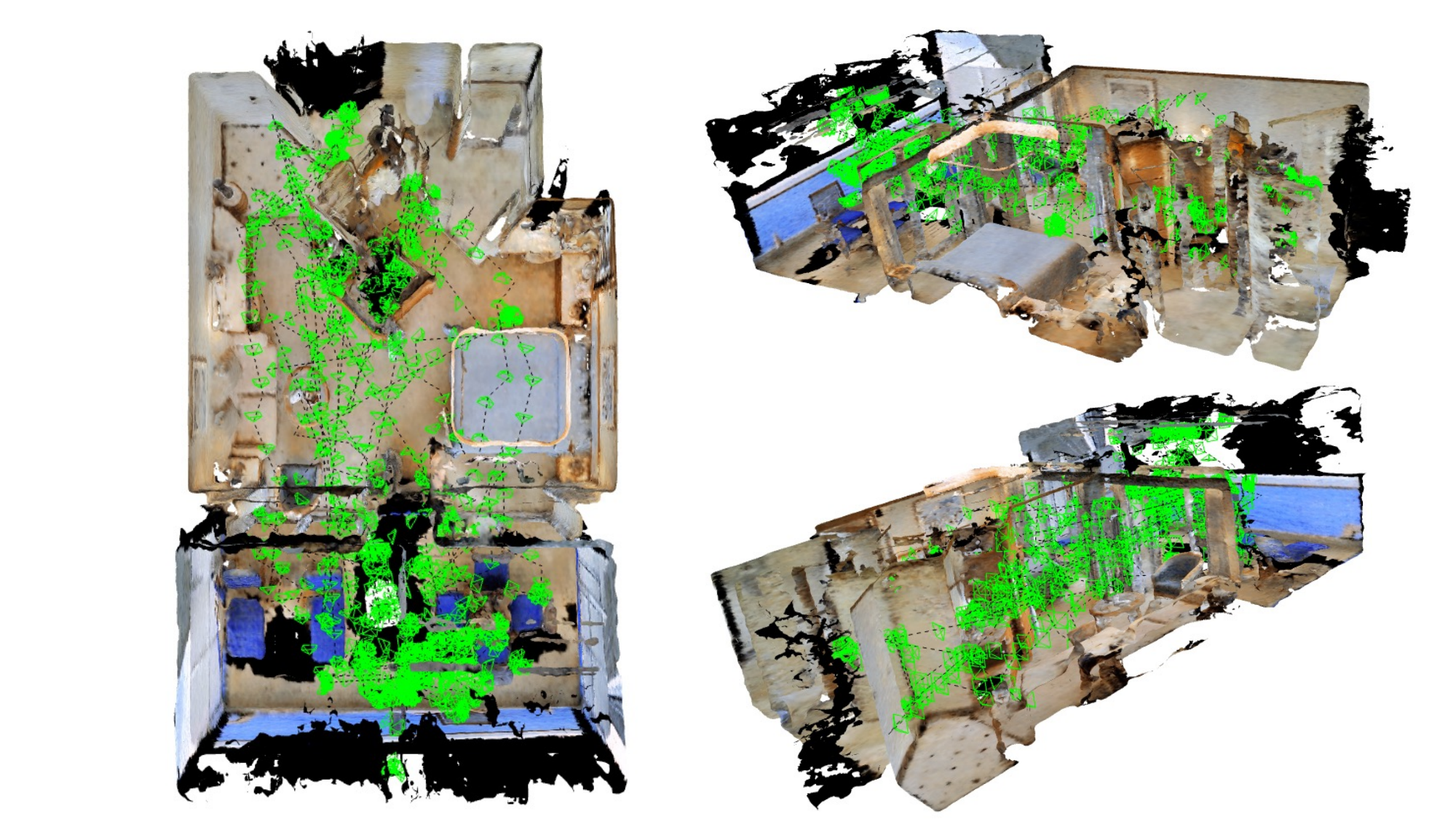}
		\caption{
            \textbf{Matterport3D (HxpK)} 
            Reconstructed Mesh and planned trajectory.
            }
            \label{fig:mp3d_HxpK}
\end{figure*}

\begin{figure*}[t!]
        \centering
		\includegraphics[width=1.\textwidth]{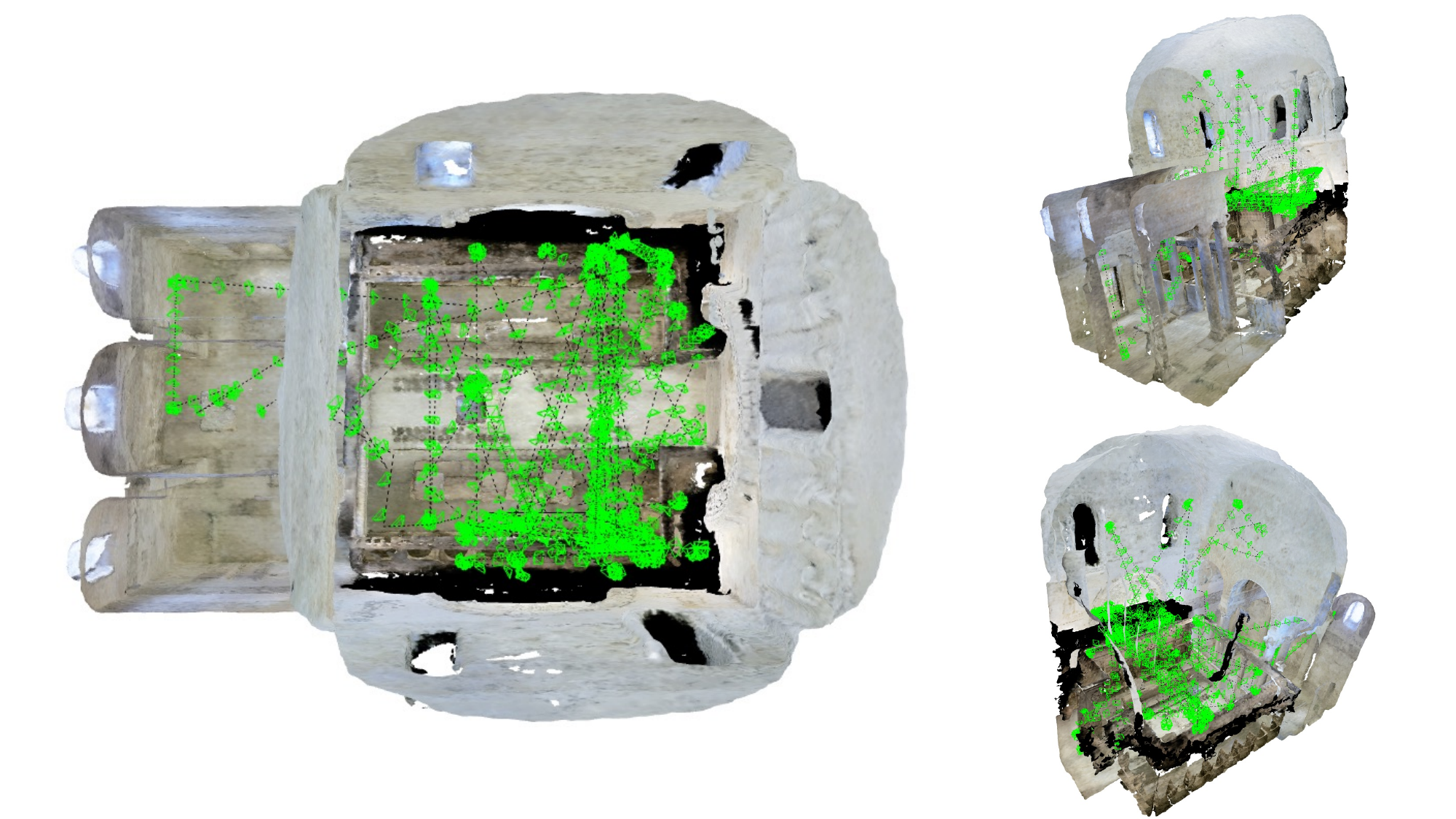}
		\caption{
            \textbf{Matterport3D (pLe4)} 
            Reconstructed Mesh and planned trajectory.
            }
            \label{fig:mp3d_pLe4}
\end{figure*}

\begin{figure*}[t!]
        \centering
		\includegraphics[width=1.\textwidth]{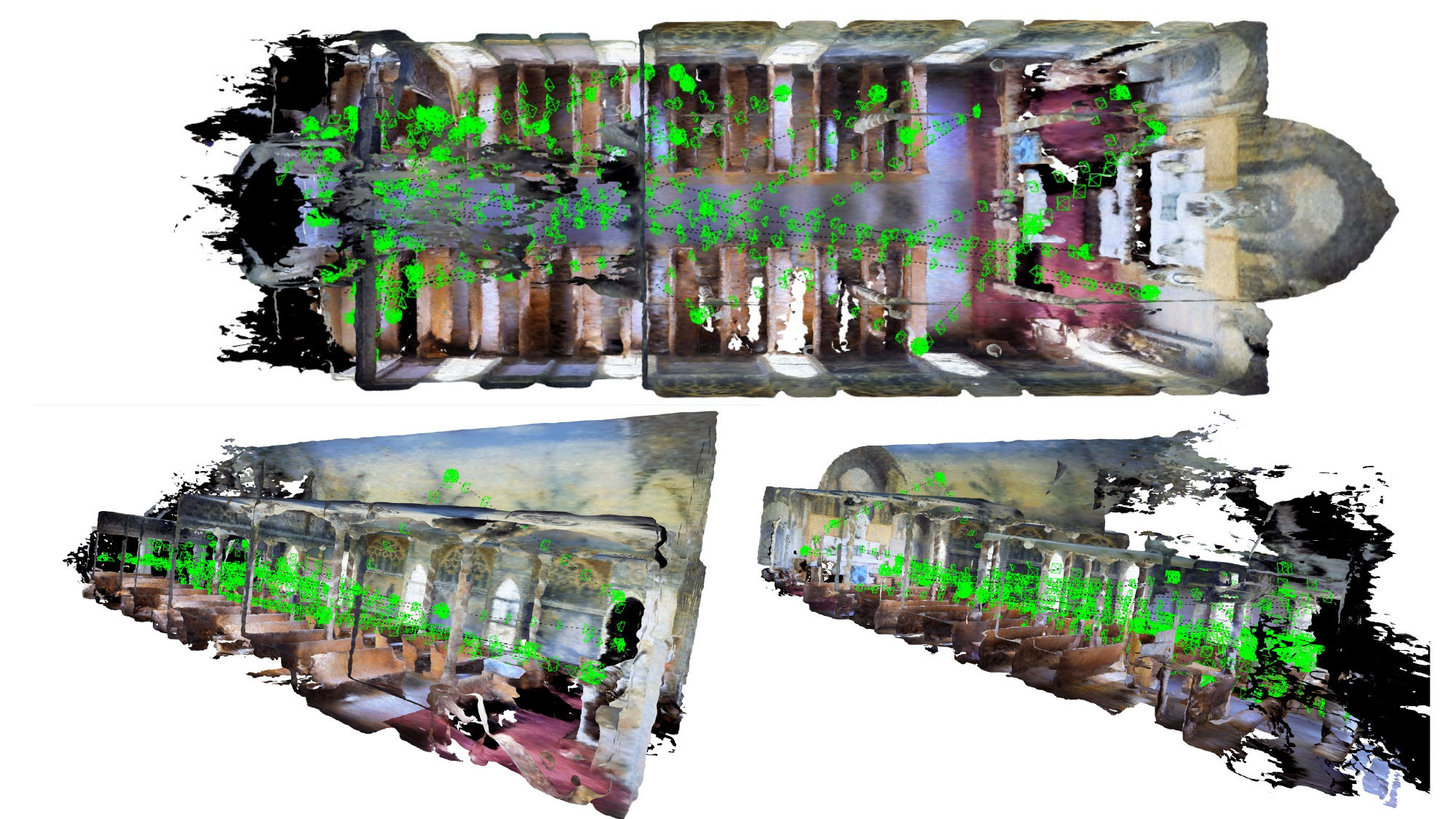}
		\caption{
            \textbf{Matterport3D (YmJk)} 
            Reconstructed Mesh and planned trajectory.
            }
            \label{fig:mp3d_YmJk}
\end{figure*}

\begin{figure*}[t!]
        \centering
		\includegraphics[width=1.\textwidth]{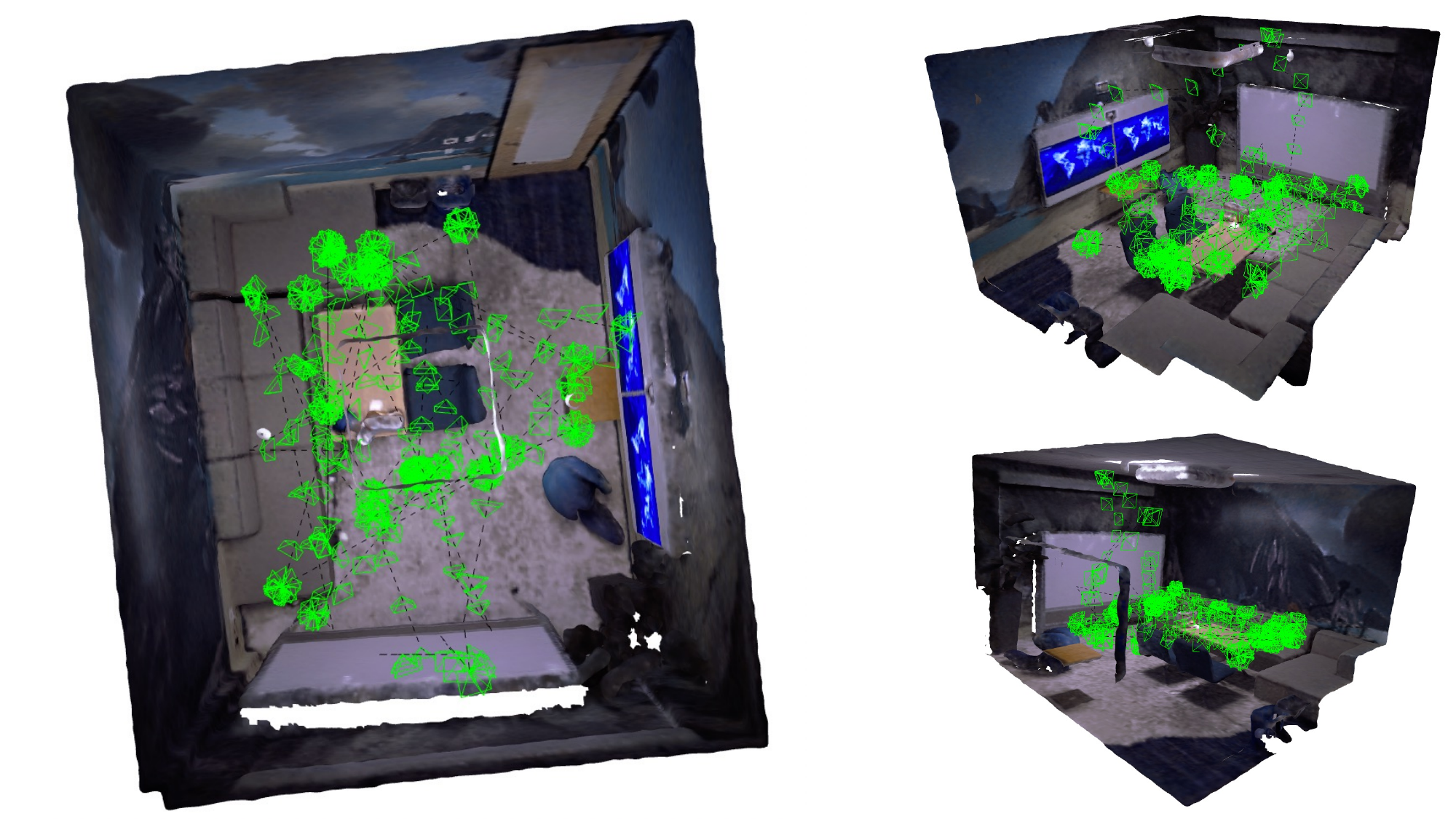}
		\caption{
            \textbf{Replica (office0)}
            Reconstructed Mesh and planned trajectory.
            }
            \label{fig:replica_office0}
\end{figure*}

\begin{figure*}[t!]
        \centering
		\includegraphics[width=1.\textwidth]{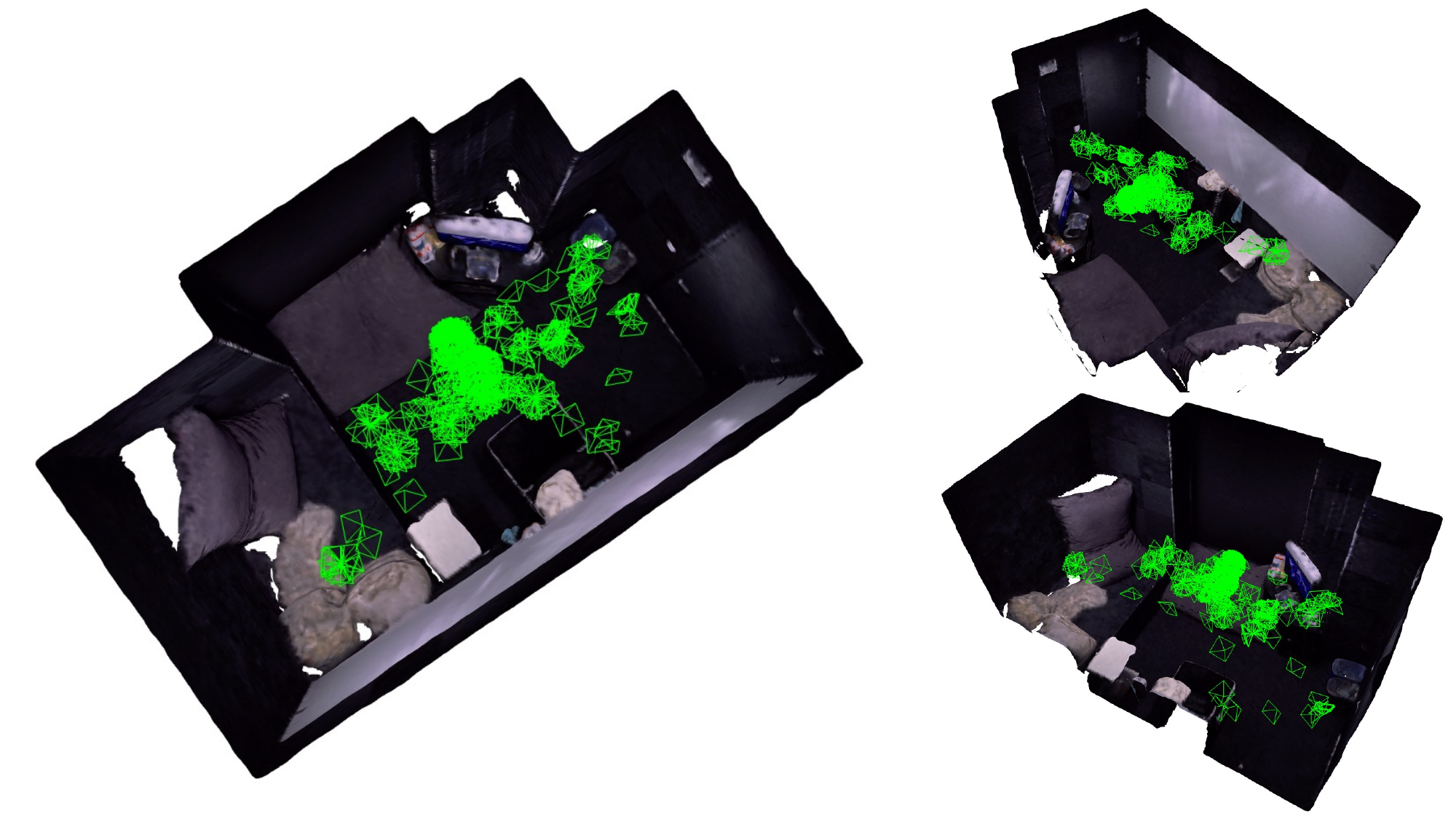}
		\caption{
            \textbf{Replica (office1)} 
            Reconstructed Mesh and planned trajectory.
            }
            \label{fig:replica_office1}
\end{figure*}

\begin{figure*}[t!]
        \centering
		\includegraphics[width=1.\textwidth]{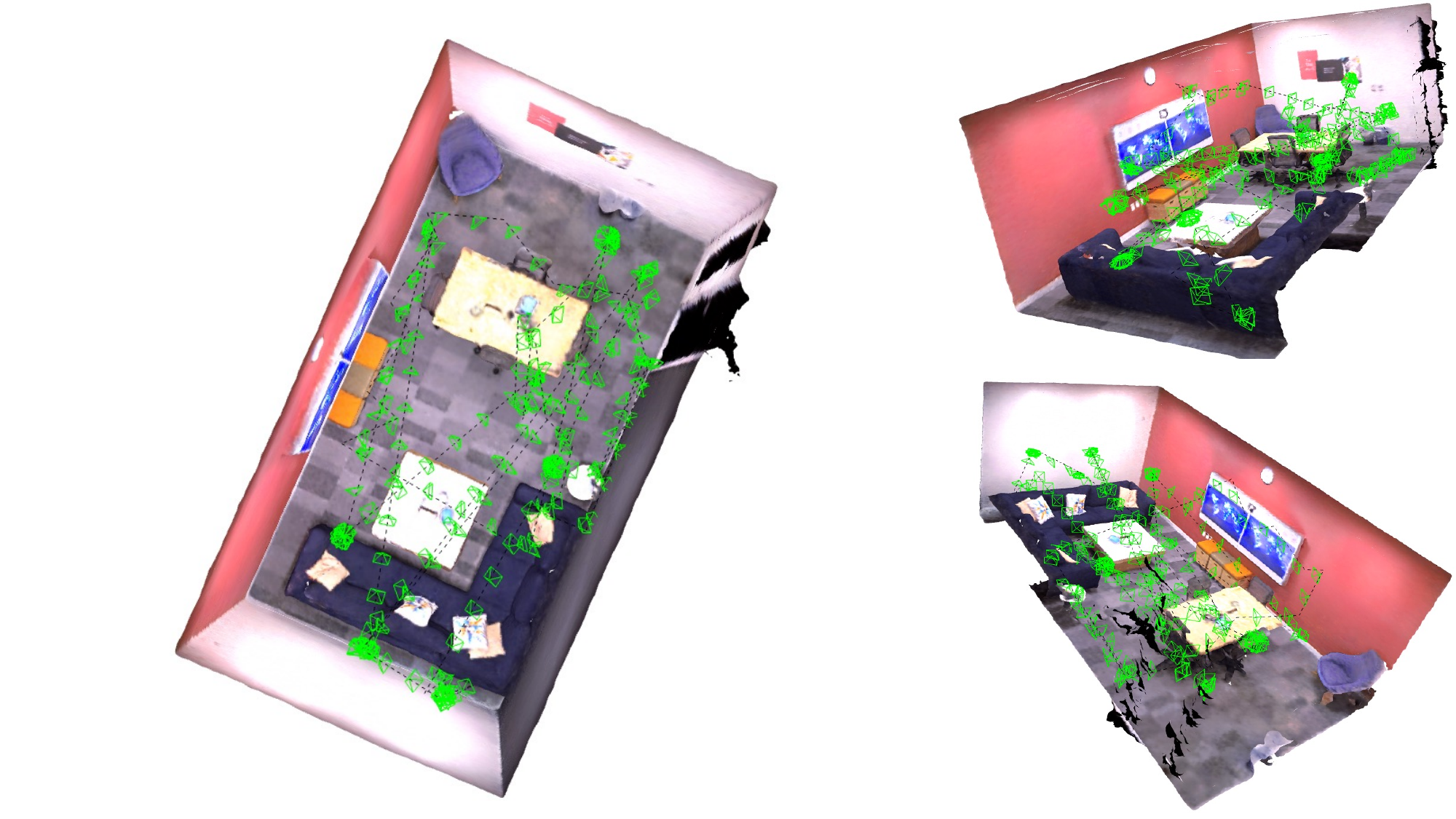}
		\caption{
            \textbf{Replica (office2)} 
            Reconstructed Mesh and planned trajectory.
            }
            \label{fig:replica_office2}
\end{figure*}

\begin{figure*}[t!]
        \centering
		\includegraphics[width=1.\textwidth]{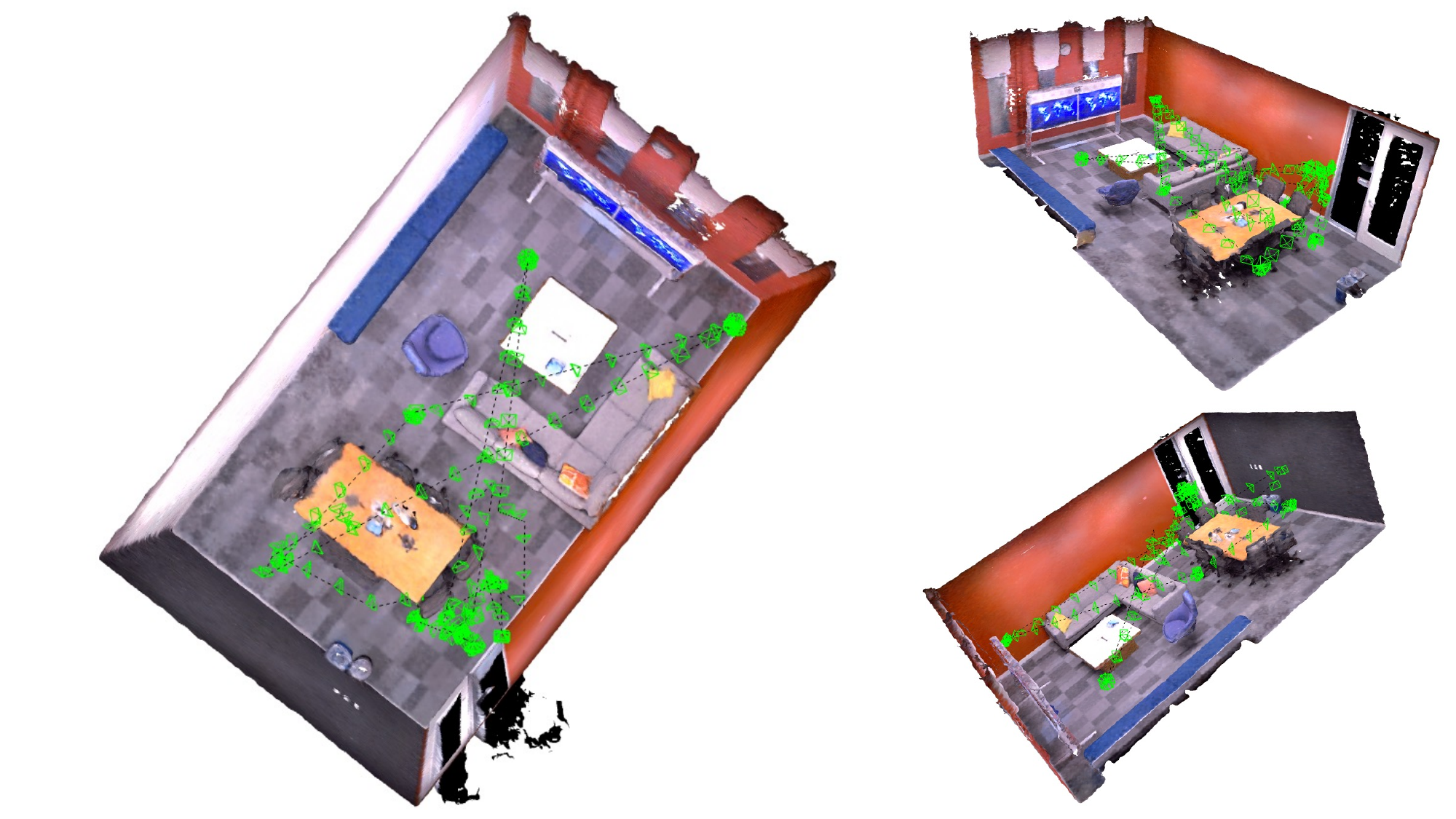}
		\caption{
            \textbf{Replica (office3)} 
            Reconstructed Mesh and planned trajectory.
            }
            \label{fig:replica_office3}
\end{figure*}

\begin{figure*}[t!]
        \centering
		\includegraphics[width=1.\textwidth]{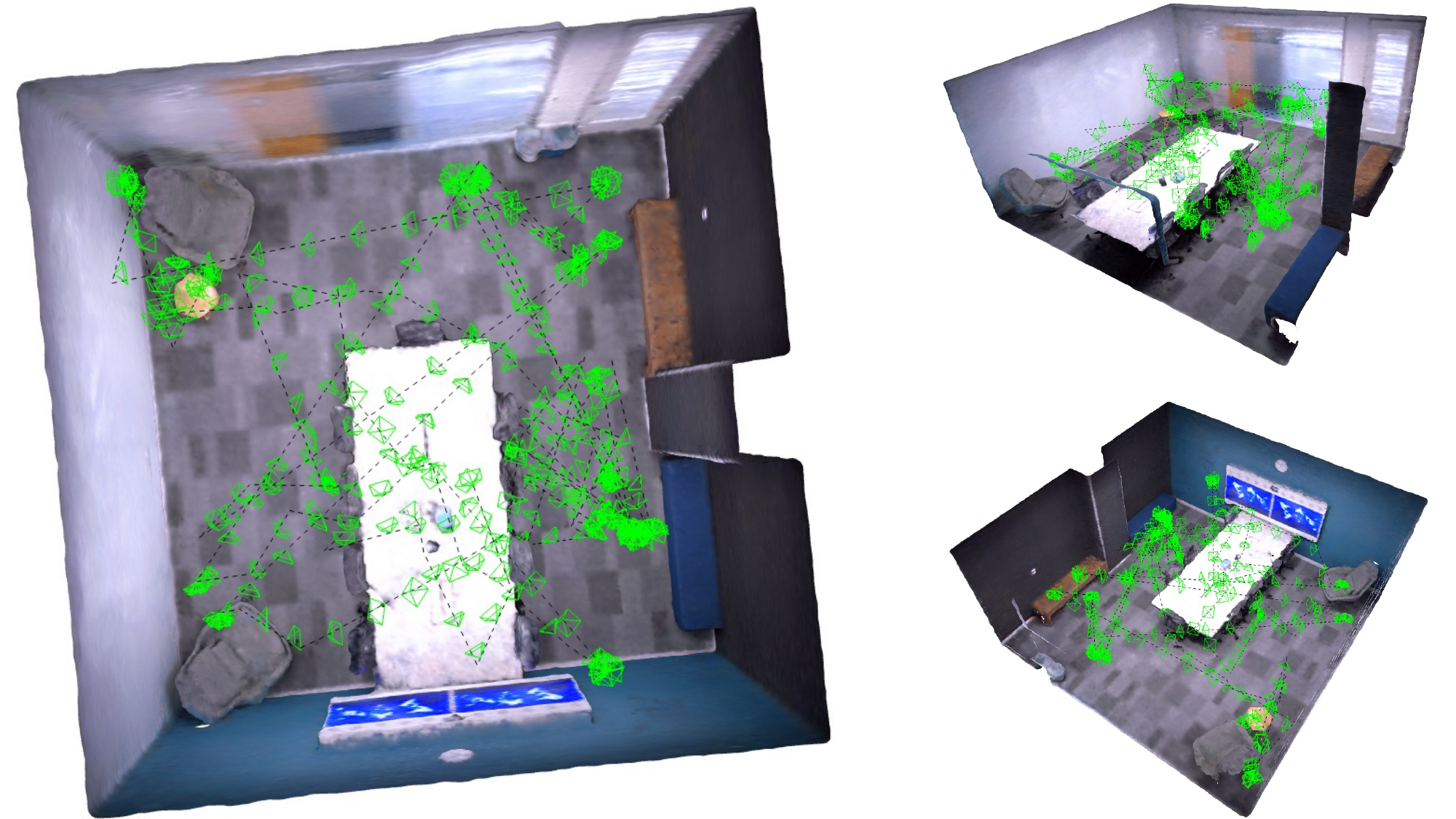}
		\caption{
            \textbf{Replica (office4)} 
            Reconstructed Mesh and planned trajectory.
            }
            \label{fig:replica_office4}
\end{figure*}

\begin{figure*}[t!]
        \centering
		\includegraphics[width=1.\textwidth]{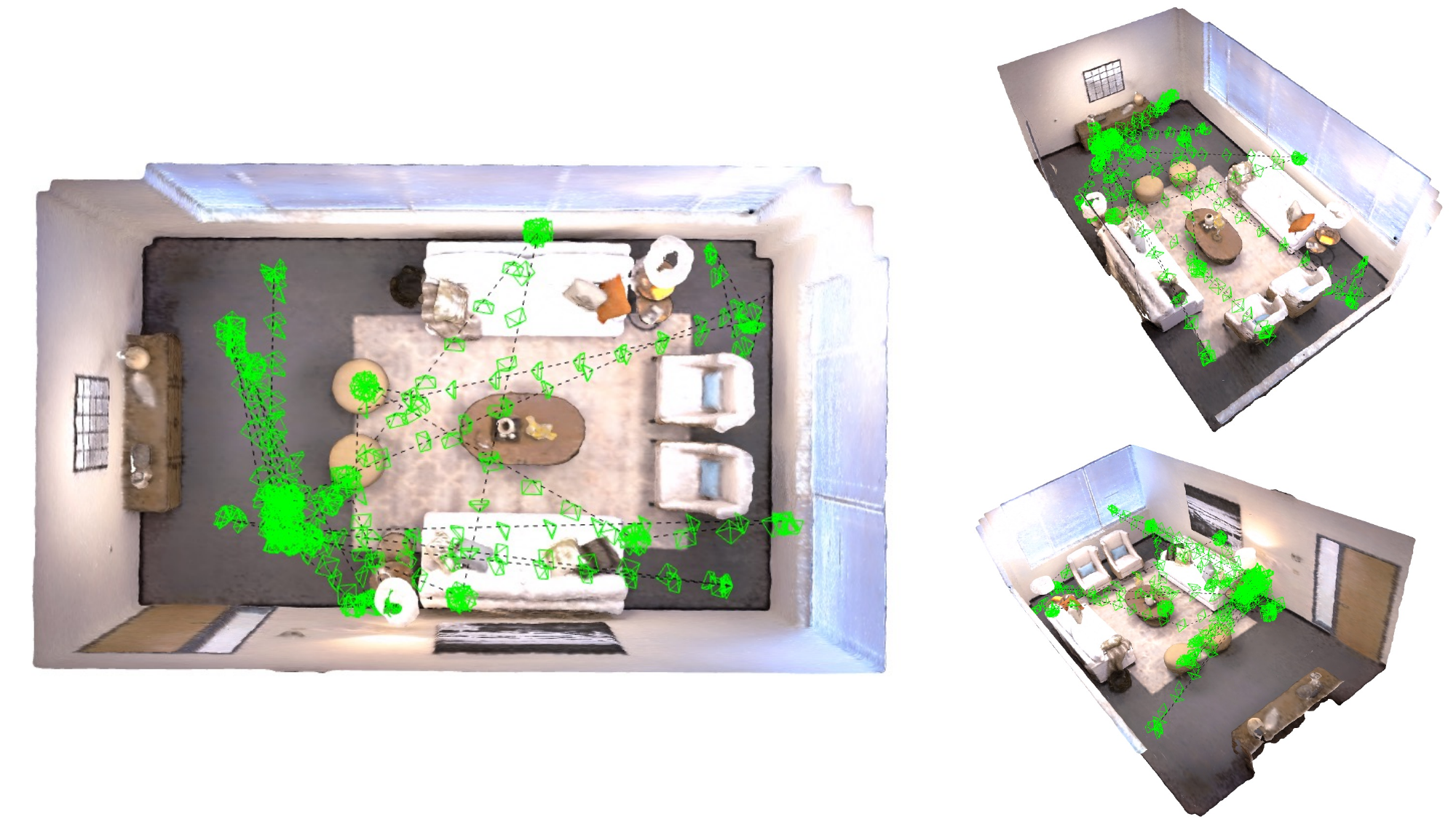}
		\caption{
            \textbf{Replica (room0)} 
            Reconstructed Mesh and planned trajectory.
            }
            \label{fig:replica_room0}
\end{figure*}

\begin{figure*}[t!]
        \centering
		\includegraphics[width=1.\textwidth]{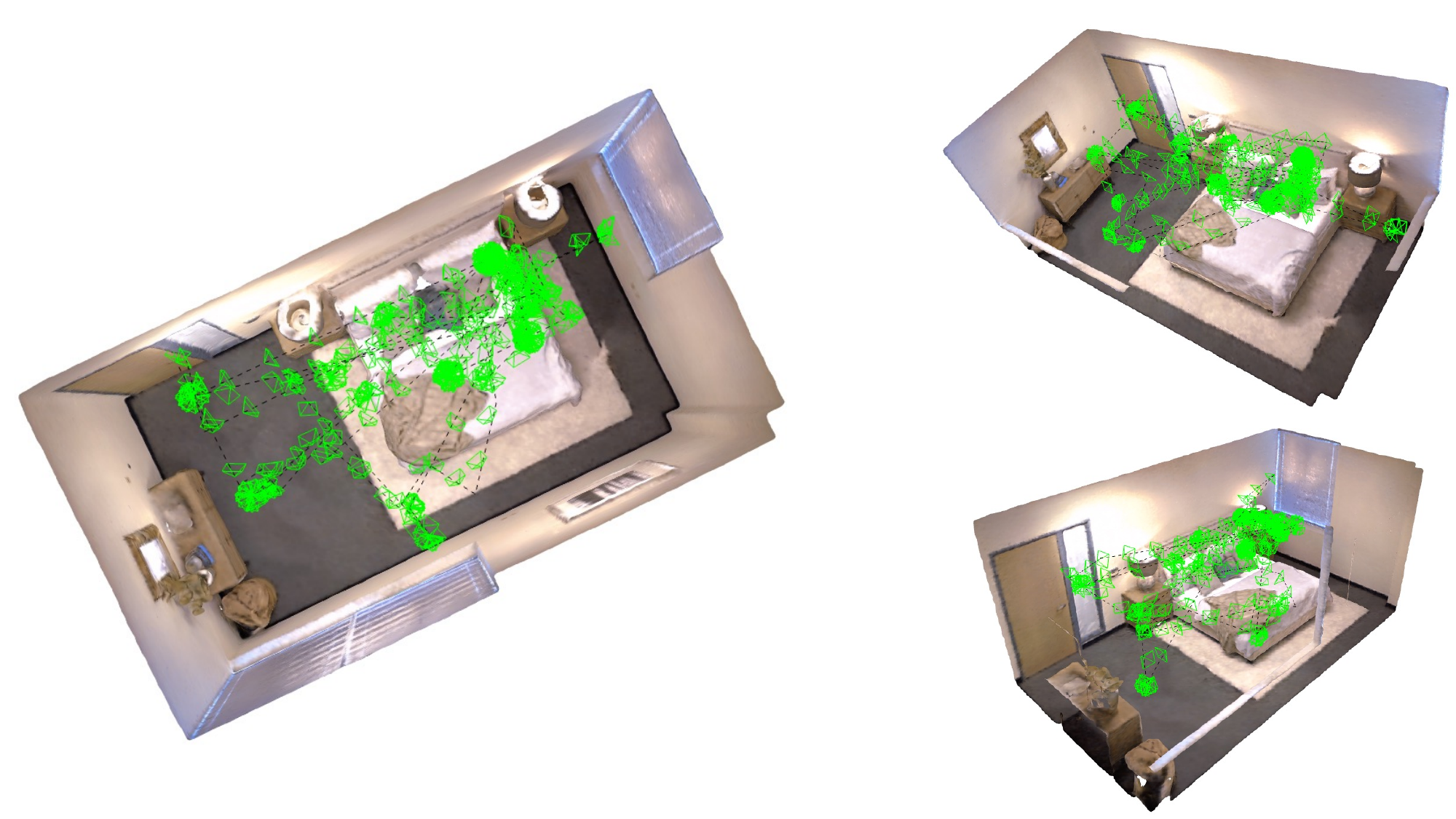}
		\caption{
            \textbf{Replica (room1)} 
            Reconstructed Mesh and planned trajectory.
            }
            \label{fig:replica_room1}
\end{figure*}

\begin{figure*}[t!]
        \centering
		\includegraphics[width=1.\textwidth]{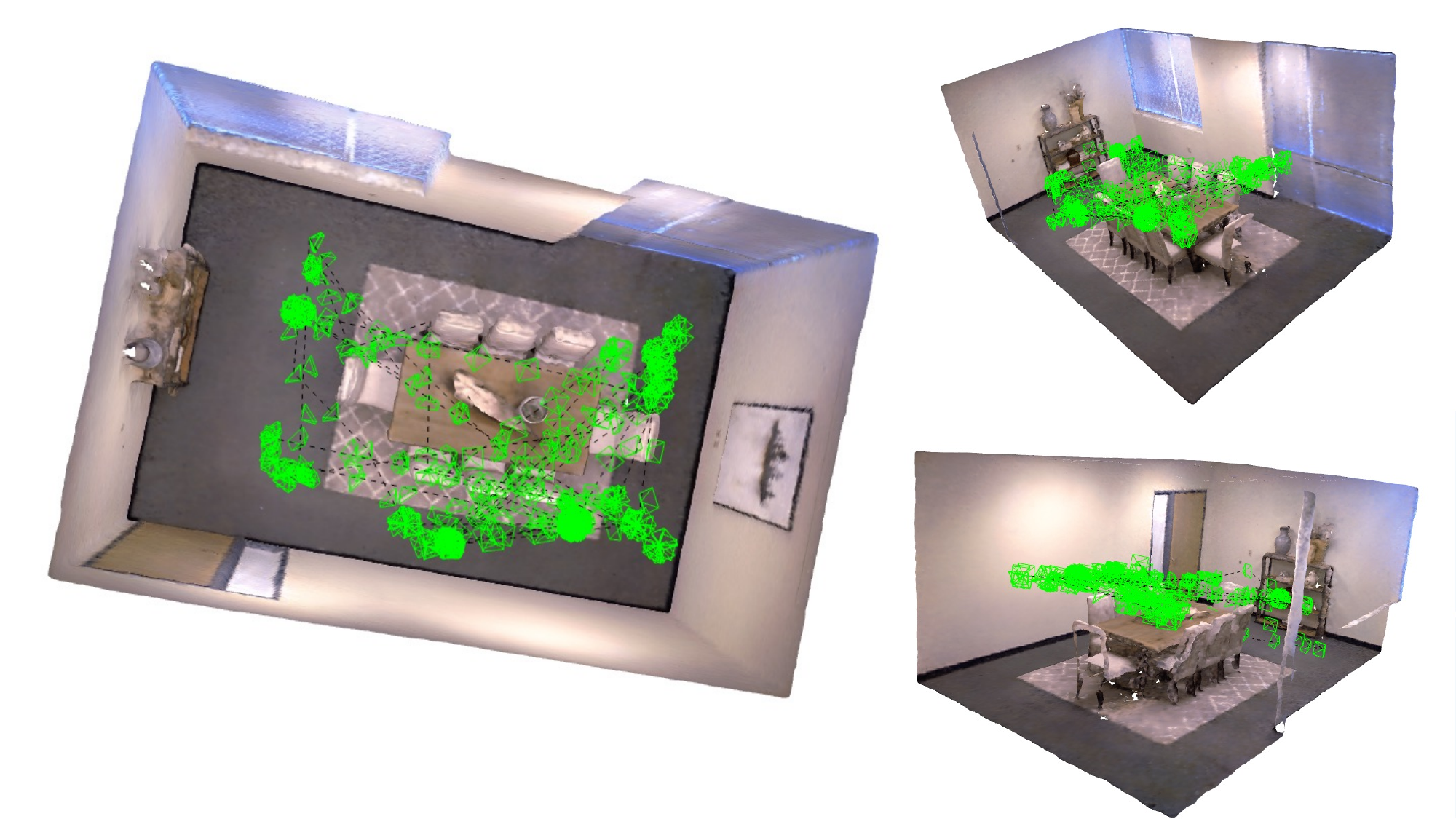}
		\caption{
            \textbf{Replica (room2)} 
            Reconstructed Mesh and planned trajectory.
            }
            \label{fig:replica_room2}
\end{figure*}

\begin{table*}[t]
    \centering
    \begin{tabular}{c|l*{8}{c}}
        \hline
        Method & Metrics & 
        office0 & office1 & office2 & office3 & office4 & room0 & room1 & room2 \\
        \hline
        \hline

            \textbf{CoSLAM}
        & Comp. Ratio $\uparrow$ & 
        96.33 & 94.65 & 92.47 & 91.43 & 91.34 & 94.67 & 95.45 & 93.95 \\
        (no tracking)
        & Traj. (m) $\uparrow$ & 
        18.20 & 11.56 & 23.16 & 29.16 & 25.22 & 24.69 & 16.21 & 23.07   \\
        \hline

        \multirow{2}{*}{
            \textbf{Ours}
        } 
        & Comp. Ratio $\uparrow$ & 
        98.17 & 95.26 & 97.54 & 93.91 & 97.93 & 98.28 & 98.04 & 98.47 \\
        & Traj. (m) $\uparrow$ & 
         81.27 & 30.02 & 90.20  & 88.59 & 96.36 & 73.91 & 96.99  & 41.31 \\
        \hline

    \end{tabular}
    \caption{Per-scene trajectory length evaluation on Replica\cite{straub2019replica} dataset}
    \label{tab:replica_traj_len}
\end{table*}


In this section, we present more comprehensive results for the various scenes included in the Matterport3D \cite{chang2017matterport3d} and Replica dataset \cite{straub2019replica}. 
Detailed, scene-specific quantitative results are provided in \cref{tab:mp3d_full} and \cref{tab:replica_full}. 
For the qualitative visualization, the reconstructed meshes undergo a culling process as delineated in Neural RGB-D \cite{azinovic2022neural} and GoSURF \cite{wang2022gosurf}, ensuring that only the most relevant data is presented. 

\paragraph{MP3D}
In \cref{tab:mp3d_full}, we present a comparative analysis of our method against the state-of-the-art Active Neural Mapping (ANM) \cite{yan2023active}. 
The results demonstrate that our method outperforms ANM across all evaluated metrics. 
Most notably, our method exhibits a significant advancement in terms of reconstruction quality and completeness, surpassing the existing benchmarks set by previous art. 
This consistent superiority in performance underscores the effectiveness of our approach in challenging reconstruction scenarios.

In \cref{fig:mp3d_result_full}, we conduct a qualitative evaluation of our 3D reconstruction method against the ground truth for various scenes in the Matterport3D dataset. 
Ground truth meshes are presented in the odd-numbered rows, while the even-numbered rows showcase our method's reconstructed meshes. 
Each scene is identified by a unique code (\eg, ``Gdvg", ``gZ6f") on the left. 
We offer a tripartite comparison for each: 
the first and second columns depict the exterior surfaces; 
the third and fourth columns reveal the interior surfaces; 
and the final two columns provide close-up views of the intricate internal reconstructions. 
This format delineates a comprehensive visual assessment, contrasting both the textural and geometric dimensions of the meshes.
 
In \cref{fig:mp3d_Gdvg} through \cref{fig:mp3d_YmJk}, we present per-scene trajectory visualizations on the Matterport3D dataset. 
For enhanced visual clarity, we focus exclusively on illustrating the trajectory formed by keyframe camera poses and the reconstructed texture mesh. 
To provide a thorough perspective of each scene, we include a bird’s eye view alongside two distinct side views. 
This tri-view presentation facilitates a comprehensive understanding of the spatial dynamics in each scene.
It is important to note that the ``black regions" visible in the mesh represent areas lacking ground truth data, which were consequently excluded from the mapping optimization process. 
Our observations indicate that while our method demonstrates high completeness in fully exploring the environment, it tends to allocate a considerable number of steps to survey these ``black regions". 
This behavior can be attributed to our selective exclusion of these regions during mapping optimization, which in turn, prevents effective reduction of uncertainty in these areas. 
Our method, prioritizing observation of uncertain regions, thus allocates more attention to these parts. 
This phenomenon is a reflection of the challenges posed by the imperfect simulation of real-world environments.

\paragraph{Replica}
We present per-scene ablation studies on Replica in \cref{tab:replica_full}.
These results demonstrate that Active Ray Sampling enhances the performance of CoSLAM \cite{wang2023coslam}, particularly in scenarios where tracking is disabled. 
Additionally, our ablation studies reveal that employing the Uncertainty Grid (Full) approach yields superior results compared to the Uncertainty Net across most scenes.

In \cref{fig:replica_result_full}, we conduct a qualitative evaluation of our 3D reconstruction method against the ground truth for various scenes in the Replica dataset. 
Ground truth meshes are presented in the odd-numbered rows, while the even-numbered rows showcase our method's reconstructed meshes. 
Our results show a high level of quality and completeness, closely mirroring the ground truths.

In \cref{fig:replica_office0} - \cref{fig:replica_room2}, we present trajectory visualization for each scene.
Given that five trials were conducted for each scene, we selectively showcase the most illustrative visualization result for demonstration purposes.
In our qualitative analysis, we present two key elements for each scene: the texture mesh visualization and the corresponding planned trajectory. 
Similarly, we only illustrate the trajectory formed by keyframe camera poses and the reconstructed texture mesh for better clarity. 

\subsection{More qualitative comparison on MP3D}

\begin{figure*}[t!]
        \centering
		\includegraphics[width=1.0\textwidth]    {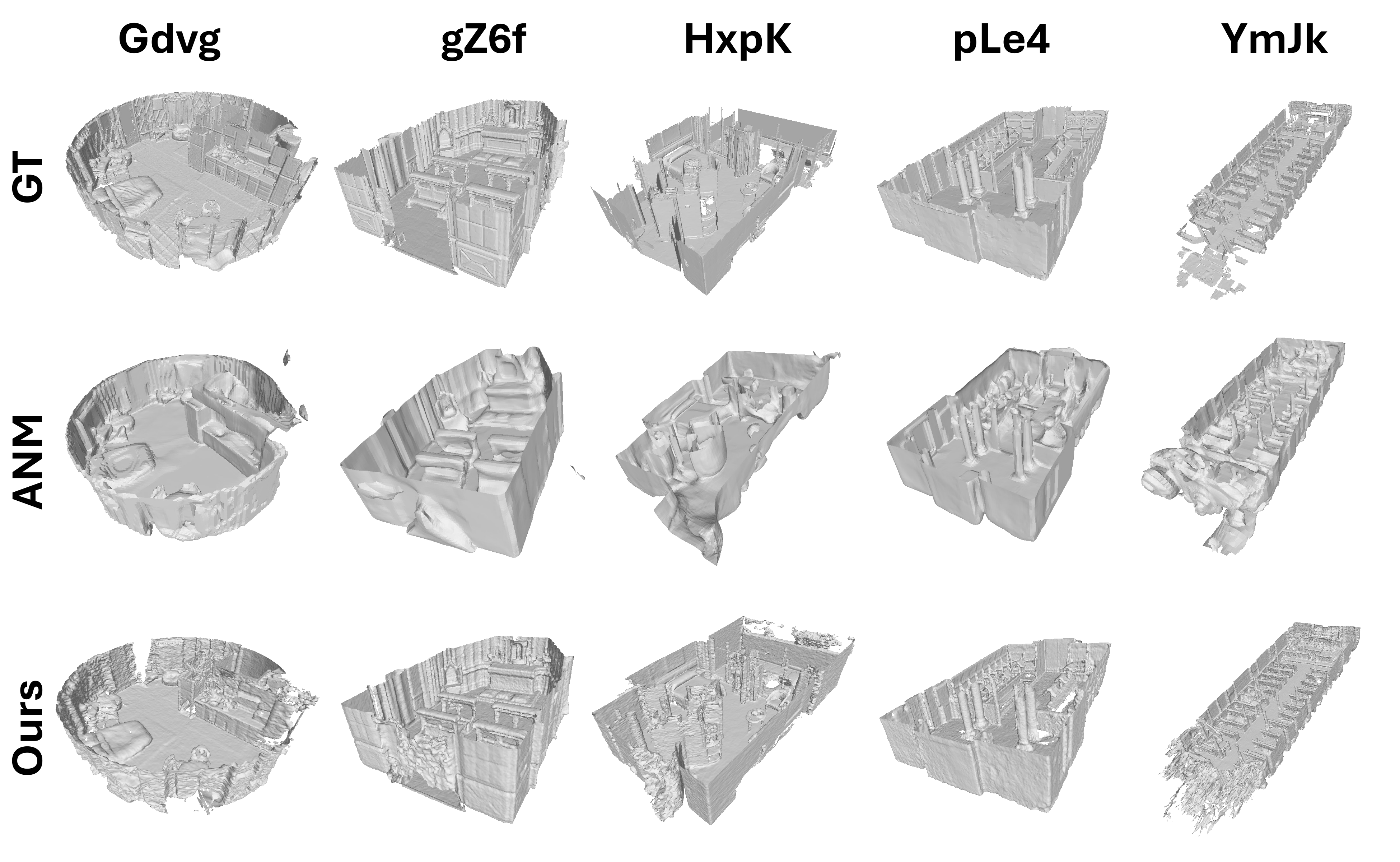}
		\caption{
           \textbf{More Matterport3D results} 
           We trim the reconstruction results for a better comparison. 
           Compared to the baseline method, ANM \cite{yan2023active}, our method shows more precise and complete reconstructions. 
            }
		\label{fig:trimmed_mp3d}
\end{figure*}
For the completeness of the study, we provide more comparison between ground truth, ANM baseline \cite{yan2023active}, and our method in Matterport3D dataset, as shown in \cref{fig:trimmed_mp3d}. 
We trim the meshes for a better visualization purpose.

\subsection{Comparison against passive mapping methods}

\begin{figure*}[t!]
        \centering
		\includegraphics[width=1.0\textwidth]    {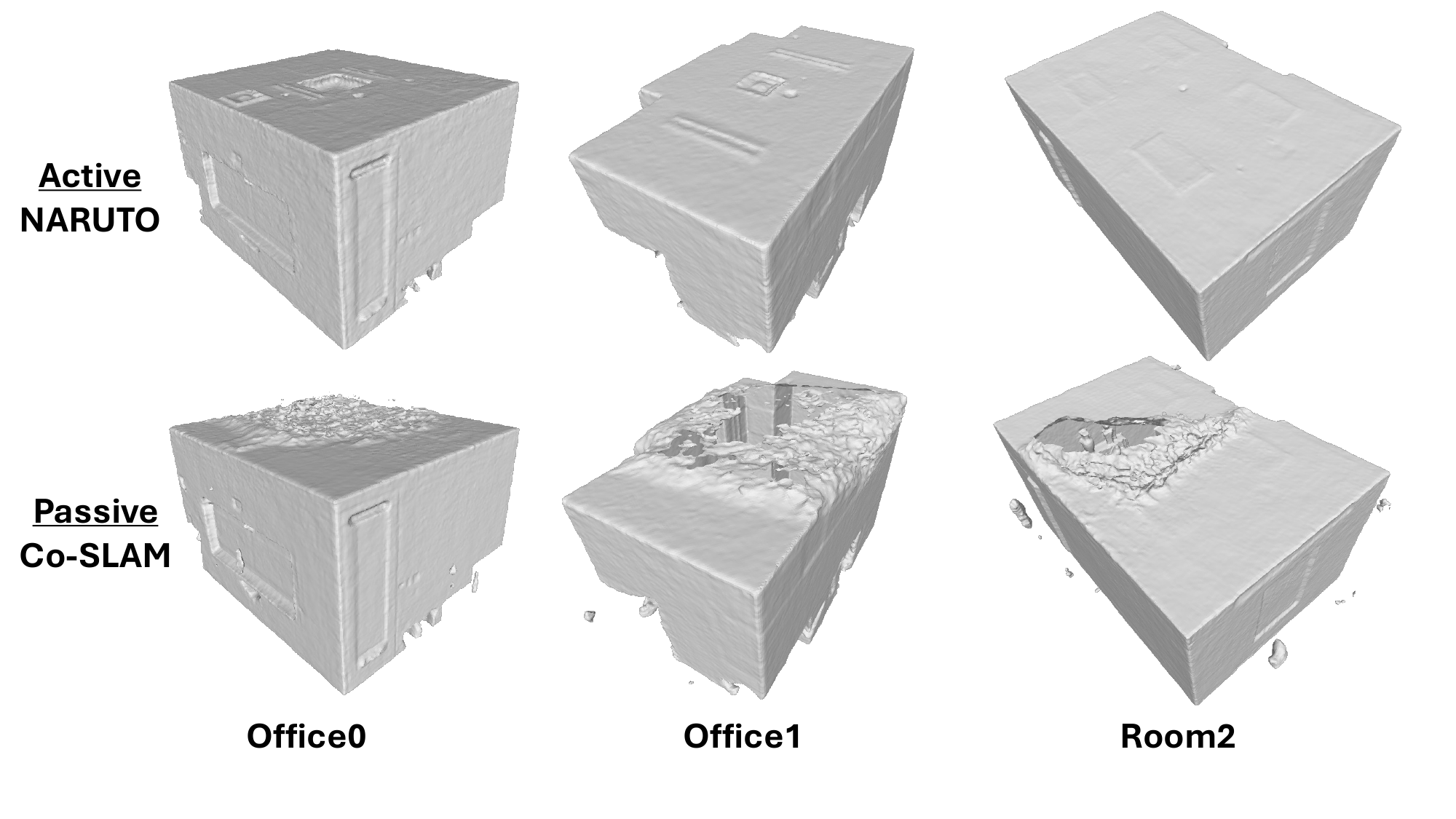}
		\caption{
           \textbf{Qualitative comparison between active and passive mapping methods. }
           For Co-SLAM \cite{wang2023coslam}, we disable the tracking thread and run the reconstruction using a pre-defined trajectory. 
           Active NARUTO shows a more complete and precise reconstruction, especially for the regions that have not been adequately covered by the passive scanning.
            }
		\label{fig:qualitative_passive}
\end{figure*}
In traditional mapping methods, typically involving environments scanned by human-operated sensing devices, the trajectory of scanning significantly impacts the reconstruction's quality and completeness. 
Such approaches are termed passive mapping methods, characterized by the absence of a planning or guidance module. 
In \cref{tab:replica_ablation}, we present a quantitative comparison between Passive Neural Mapping and Active Neural Mapping, utilizing Co-SLAM as the backbone. 
Here, we aim to offer additional qualitative comparisons in \cref{fig:qualitative_passive} to highlight differences in reconstruction details more vividly.
In passive Co-SLAM (with tracking disabled), regions may be missed or poorly reconstructed if not adequately covered by the scanning trajectory. 
Conversely, our active reconstruction method ensures a more comprehensive and accurate reconstruction, effectively addressing these limitations.

We compared the trajectory lengths of passive versus active scanning on the Replica dataset, with the results detailed in \cref{tab:replica_traj_len}. 
Under the same conditions (2000 frames with 400 keyframes), passive scanning may result in redundant observations due to the lack of guided scanning. 
Active scanning, on the other hand, enables more extensive coverage and yields superior reconstruction quality. 
However, this approach typically results in longer trajectories, as the agent continuously moves to ensure comprehensive scanning of the environment.







\end{document}